\documentclass{egpubl}
\usepackage{eurovis2020}

\usepackage[absolute,overlay]{textpos}
\usepackage{pbox}
\usepackage{everypage}

\newcommand{\stamp}[1][© 2020 The Eurographics Association and John Wiley \& Sons Ltd. This is the author's version of the article that has been published in Computer Graphics Forum. The final version of this record is available at: \href{https://doi.org/10.1111/cgf.14034}{\color{blue}10.1111/cgf.14034}]{%
\begin{textblock*}{140mm}(37mm,270mm)
\centering%
\small%
\emph{#1}%
\end{textblock*}%
}

\AddEverypageHook{
  \stamp
}

 \STAREurovis   
\usepackage[T1]{fontenc}
\usepackage{dfadobe}  

\usepackage[normalem]{ulem}

\usepackage[dvipsnames]{xcolor} 

\newcommand{\circled}[1]{\raisebox{.5pt}{\textcircled{\raisebox{-.9pt} {#1}}}}
\newcommand{\circledTwo}[1]{\raisebox{.5pt}{\textcircled{\raisebox{-.2pt} {#1}}}}

\newcommand{\cit}[1]{``#1''}

\usepackage{cite}  
\BibtexOrBiblatex
\electronicVersion
\PrintedOrElectronic
\ifpdf \usepackage[pdftex]{graphicx} \pdfcompresslevel=9
\else \usepackage[dvips]{graphicx} \fi
\graphicspath{{figures/}{pictures/}{images/}{./}} 

\usepackage{cleveref}
\renewcommand{\autoref}{\Cref}

\usepackage{egweblnk}

\newcommand{\statbox}[2]{\noindent\raisebox{0.7\height}{\fcolorbox[rgb]{0,0,0}{#1}{\scriptsize~}}}

\usepackage{longfbox}

\newfboxstyle{tight2}{padding=2pt,margin=0pt,baseline-skip=false}%

\definecolor{ColT1}{HTML}{1f77b4}
\definecolor{ColT2}{HTML}{ff7f0e}
\definecolor{ColT3}{HTML}{2ca02c}
\definecolor{ColT4}{HTML}{d62728}
\definecolor{ColT5}{HTML}{9467bd}
\definecolor{ColT6}{HTML}{8c564b}
\definecolor{ColT7}{HTML}{e377c2}
\definecolor{ColT8}{HTML}{7f7f7f}
\definecolor{ColT9}{HTML}{bcbd22}
\definecolor{ColT10}{HTML}{17becf}

\newlength{\boxh}
\title[Enhancing Trust in Machine Learning Models with the Use of Visualizations]%
      {The State of the Art in Enhancing Trust in Machine Learning Models with the Use of Visualizations}

\author[Chatzimparmpas et al.]
{\parbox{\textwidth}{\centering A. Chatzimparmpas$^{1}$\orcid{0000-0002-9079-2376}, R.\,M. Martins$^{1}$\orcid{0000-0002-2901-935X}, I. Jusufi$^{1}$\orcid{0000-0001-6745-4398}, K. Kucher$^{1}$\orcid{0000-0002-1907-7820}, F. Rossi$^{2}$\orcid{0000-0003-4638-1286}, and A. Kerren$^{1}$\orcid{0000-0002-0519-2537}}
        \\
{\parbox{\textwidth}{\centering $^1$Department of Computer Science and Media Technology, Linnaeus University, Sweden\\ $^2$Ceremade, Universit{\'{e}} Paris Dauphine, PSL University, France}}
}

\begin{document}

\stamp

\maketitle

\begin{abstract}
   Machine learning (ML) models are nowadays used in complex applications in various domains, such as medicine, bioinformatics, and other sciences. Due to their black box nature, however, it may sometimes be hard to understand and trust the results they provide. This has increased the demand for reliable visualization tools related to enhancing trust in ML models, which has become a prominent topic of research in the visualization community over the past decades. To provide an overview and present the frontiers of current research on the topic, we present a State-of-the-Art Report (STAR) on enhancing trust in ML models with the use of interactive visualization. We define and describe the background of the topic, introduce a categorization for visualization techniques that aim to accomplish this goal, and discuss insights and opportunities for future research directions. Among our contributions is a categorization of trust against different facets of interactive ML, expanded and improved from previous research. Our results are investigated from different analytical perspectives: (a) providing a statistical overview, (b) summarizing key findings, (c) performing topic analyses, and (d) exploring the data sets used in the individual papers, all with the support of an interactive web-based survey browser. We intend this survey to be beneficial for visualization researchers whose interests involve making ML models more trustworthy, as well as researchers and practitioners from other disciplines in their search for effective visualization techniques suitable for solving their tasks with confidence and conveying meaning to their data.

   \makeatletter
	\def\customkeywords{\vskip 5.5pt\par\reset@font\rmfamily}
	\def\endcustomkeywords{\relax}

	\begin{customkeywords}
	\end{customkeywords}

	\def\customclassification{\vskip 5.5pt\par\reset@font\rmfamily}
	\def\endcustomclassification{\relax}
	\makeatother

	\begin{customclassification}
		\textbf{ACM CCS:} $\bullet$ Information systems $\rightarrow$ Trust;
		$\bullet$ Human-centered computing $\rightarrow$ Visual analytics;
		$\bullet$ Human-centered computing $\rightarrow$ Information visualization;
		$\bullet$ Human-centered computing $\rightarrow$ Visualization systems and tools;
		$\bullet$ Machine learning $\rightarrow$ Supervised learning;
		$\bullet$ Machine learning $\rightarrow$ Unsupervised learning;
		$\bullet$ Machine learning $\rightarrow$ Semi-supervised learning;
		$\bullet$ Machine learning $\rightarrow$ Reinforcement learning
	\end{customclassification}
\end{abstract} 

\section{Introduction} \label{sec:intro}
	Trust in machine learning (ML) models is one of the greatest challenges in real-life applications of ML~\cite{Toreini2020The}.
ML models are now commonplace in many research and application domains, and they are frequently used in scenarios of complex and critical decision-making~\cite{Nieto2019Usage,Phillips2006Artificial,Thammachantuek2018Comparison}.
Medicine, for example, is one of the fields where the use of ML might offer potential improvements and solutions to many difficult problems~\cite{Kelly2019Key,Sidey2019Machine,Shah2019Artificial}. 
A significant challenge that remains, however, is how \emph{trustworthy} are the ML models that are being used in these disciplines.
Rudin and Ustun~\cite{Rudin2018Optimized}, for example, emphasize the importance of trust for ML models in healthcare and criminal justice, since they play a significant role in making decisions regarding human lives.
It is not uncommon to observe that domain experts may not rely on ML models if they do not understand how they work~\cite{Janik2019Interpreting}. 
\\

The impact of this problem can already be observed in recent works, such as the program ``Explainable AI (XAI)'' founded by DARPA (Defense Advanced Research Projects Agency)\cite{DARPA_XAI} and described by Krause et al.~\cite{Krause2017AWorkflow}. 
This initiative is only one of the various projects that suggest further research into the field of XAI, which---to a certain extent---addresses challenges related to trust. 
The XAI program in its two main motivational points mentions specifically that ``producing more explainable models, while maintaining a high level of learning performance'' and ``enabling human users to understand, appropriately trust, and effectively manage the emerging generation of AI'' are both key actions for the future development in numerous domains that use ML. Understanding and trusting ML models is also arguably mandatory under the General Data Protection Regulation (GDPR) \cite{GDPR} as part of the ``right to be informed'' principle: data controllers must provide meaningful information about the logic involved in automated decisions \cite{WP251}. Individuals have also the right not to be subject to a decision based solely on automated processing: enabling subjects of ML algorithms to trust their decision is probably the easiest way to reduce the objection to such automated decisions.

In reaction to these aforementioned challenges, multiple new solutions have recently been proposed both in academia and in industry. Google's Explainable Artificial Intelligence (AI) Cloud\cite{GoogleCloud}, for example, assists in the development of interpretable and explainable ML models and supports their deployment with increased confidence. Another example is the Descriptive mAchine Learning EXplanations (DALEX)\cite{DALEX} package, which offers various functionalities that help users understand how complex models work. 
Some works propose to enable domain experts to collaborate with each other to tackle this problem together~\cite{Cai2019The,Feng2019Whart}. 
In this context, information visualization (InfoVis) techniques have been shown to be effective in making analysts more comfortable with ML solutions. Krause et al.~\cite{Krause2014INFUSE}, for example, present a case study of domain experts using their tool to explore predictive models in electronic health records. 
Also, in visual analytics (VA), the first stages to partially address those challenges have already been reached, for instance by discussing how global~\cite{Ribeiro2016Model} or local~\cite{Muhlbacher2014Opening} interpretability can assist in the interpretation and explanation of ML~\cite{Gilpin2018Explaining,Wolf2019Explainability}, and how to interactively combine visualizations with ML in order to better trust the underlying models~\cite{Sacha2016The}.

We build our state-of-the-art report (STAR) upon the results of existing visualization research, which has emphasized the need for improved trust in areas, such as VA in general, dimensionality reduction (DR), and data mining. 
Sacha et al.~\cite{Sacha2016The} aimed to clarify the role of uncertainty awareness in VA and its impact on human trust. They suggested that the analyst needs to trust the outcomes in order to achieve progress in the field.
Sedlmair et al.~\cite{Sedlmair2012Dimensionality} found important gaps between the needs of DR users and the functionalities provided by available methods. Such limitations reduce the trust that users can put in visual inferences made using scatterplots built from DR techniques.
Bertini and Lalanne~\cite{Bertini2009Surveying} concluded, from a survey, that visualization can improve model interpretation and trust-building in ML.
An interesting paper by Ribeiro et al.~\cite{Ribeiro2016Why} shows that the interest on using visualization to handle issues of trust is also present in the ML field. The authors describe a method that explains the predictions of any classifier via textual or visual cues, providing a qualitative understanding of the relationship between the instance's components.
Despite all the currently proposed solutions, many unanswered questions and challenges still remain, e.g., 
(1) If the analysts are not aware of the inherent uncertainties and trust issues that exist in an ML system, how to ensure that they do not form wrong assumptions? 
(2) Are there any guarantees that they will not be deceived by false (or unclear) results? (3) What problems of trustworthiness arise in each of the phases of a typical ML pipeline?

In this STAR, we present a general mapping of the currently available literature on using visualization to enhance trust in ML models. The mapping consists of details about which visualization techniques are used, what their reported effectiveness levels are, which domains and application areas they apply to, a conceptual understanding of what trust means in relation to ML models, and what important challenges are still open for research. Note that the terms \emph{trust} and \emph{trustworthiness} are used interchangeably throughout the report. 
The main scientific contributions of this STAR are:

\begin{itemize}
    \item an empirically informed definition of what trust in ML models means;
    \item a fine-grained categorization of trust against different facets of interactive ML, extracted from 200 papers from the past 12 years; 
    \item an investigation of existing trends and correlations between categories based on temporal, topic, and correlation analyses;
    \item the deployment of an interactive online browser (see below) to assist researchers in exploring the literature of the area; and 
    \item further recommendations for future research in visualization for increasing the trustworthiness of ML models.
\end{itemize}

\noindent To improve our categorization, identify exciting patterns, and promote data investigation by the readers of this report, we have deployed an interactive online survey browser available at
\begin{center}
\url{https://trustmlvis.lnu.se}
\end{center}
We expect that our results will support new research possibilities for different groups of professionals:

\begin{itemize}
    \item beginners/non-experts who want to get acquainted with the field quickly and gain trust in their ML models;
    \item domain experts/practitioners of any discipline who want to find the appropriate visualization techniques to enhance trust in ML models;
    \item model developers and ML experts who investigate techniques to boost their confidence and trust in ML algorithms and models; and
    \item early-stage and senior visualization researchers who intend to develop new tools and are in search of motivation and ideas from previous work.
\end{itemize}

\begin{figure*}[ht]
 \centering 
\includegraphics[width=\textwidth]{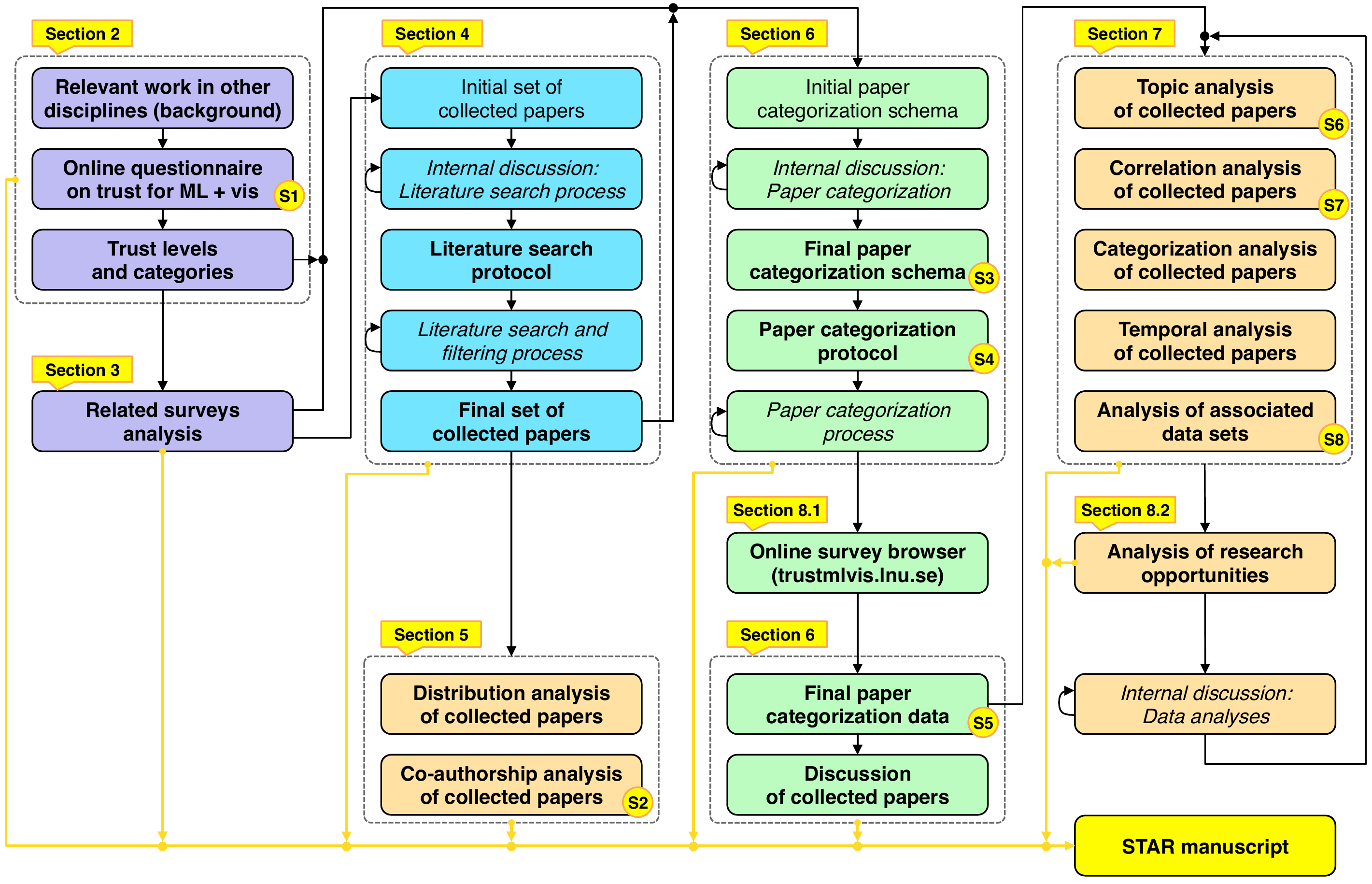}
 \caption{The overview of our STAR with regard to the methodology, main results, and corresponding sections of the manuscript. 
 Color coding is used for grouping related activities and results (purple for the background information and key concepts, blue for the literature search, green for the paper categorization, orange for the data analyses, and yellow for the manuscript); italic font is used for intermediate activities; and bold font is used for the items discussed explicitly in this STAR. The marks \circledTwo{\tiny S1}--\circledTwo{\tiny S8} refer to supplementary materials.}
 \label{fig:star-overview}
 \vspace{-.9em}
\end{figure*}

\noindent The rest of this report is organized as follows (see \autoref{fig:star-overview}). In \autoref{sec:back}, we introduce background information that we used in order to comprehend the concept of trustworthiness of ML models. We also describe our adopted definition of the meaning of trust in ML models. In \autoref{sec:relwo}, we discuss existing visualization surveys that are relevant to our work. Afterwards, \autoref{sec:meth} provides details with regard to our methodology, i.e., the searched venues and the paper collection process. The overview in \autoref{sec:overview} includes initial statistical information. 
In \autoref{sec:categ}, we present our categorization and describe the most representative examples. In \autoref{sec:topic}, we report the results of a topic analysis performed on these papers to find new and interesting topics and trends derived from them, and further findings from data-driven analysis. Our interactive survey browser and research opportunities are discussed in \autoref{sec:discuss}. Finally, \autoref{sec:conclusion} concludes the STAR.
Additionally, a set of supplementary materials (referred to as \textbf{S1} to \textbf{S8}) is also available, including the documents used to guide our categorization methodology, as well as the data that could not be part of this report due to space restrictions. 

\section{Background: Levels of Trustworthiness of Machine Learning Models} \label{sec:back}
	First, we present some earlier definitions of trust that are subsequently adapted to the context of our research. We also discuss qualitative data gathered from an online questionnaire that we distributed among ML experts and practitioners. The goals of the questionnaire were to shape our categorization of trust issues in ML and to bring to light potential ideas on how visualization can support the improvement of trustworthiness in the ML process. Building upon these definitions and results, we group the identified factors of trust into five \textbf{\emph{trust levels} (TLs)}. These levels are a part of our overall methodology, discussed in \autoref{sec:categ}.

\noindent \textbf{Definitions of trust.} \quad
The issues of definition and operationalization of \emph{trust} have been discussed in multiple research disciplines, including psychology~\cite{Evans2009The} and management~\cite{Mayer1995An}. 
Such definitions typically focus on trust in the context of expectations and interactions between individuals and organizations. 
The existing work in human-computer interaction (HCI) extends this perspective. For example, Shneiderman~\cite{Shneiderman2000Designing} provides guidelines for software development that should facilitate the establishment of trust between people and organizations. To ensure \emph{trustworthiness} of software systems, he recommends the involvement of independent oversight structures~\cite{Shneiderman2020Human}. 
Fogg and Tseng~\cite{Fogg1999The} state that \cit{trust indicates a positive belief about the perceived reliability of, dependability of, and confidence in a person, object, or process}; in their work, trust is also related (and compared) to the concept of \emph{credibility}. 
Rather than focusing on \emph{interpersonal} trust, the existing work has also addressed \emph{trust in automation}~\cite{Hoffman2013Trust}, which is more relevant to our research problem. 
Lee and See provide the following definition, widely used by the researchers in this context~\cite{Lee2004Trust}: trust is \cit{the attitude that an agent will help achieve an individual's goals in a situation characterized by uncertainty and vulnerability}. 
This definition has been further extended by Hoff and Bashir~\cite{Hoff2015Trust}, who propose a model of trust in automation with factors categorized into multiple \emph{dimensions} and \emph{layers}.  
Further adaptation of such multi-dimensional approach has been demonstrated, for example, by Yu et al.~\cite{Yu2018Do}. 
Lyons et al.~\cite{Lyons2018Trust} adopt a model consisting of a non-orthogonal set of factors in their analysis of trust factors for ML systems. 
\textbf{In this STAR}, we rely on the rather general definition of trust by Lee and See~\cite{Lee2004Trust} and further expand it into a more detailed, multi-level model presented below. 
Additionally, we make use of the definitions and factors of trust described in the existing work within InfoVis and VA and incorporate them in our model. 
For example, Chuang et al.~\cite{Chuang2012Interpretation} define trust as ``the actual and perceived accuracy of an analyst's inferences''. 
Although important, this definition touches only on a single perspective of trust: the one related to the users' expectations.
The authors also mention that usually, during evaluations of the design choices performed on new visualization systems and tools, the modelling of choices and relationships \emph{between} views is often omitted. This practice introduces limitations regarding the improvement of trust for the system as a whole, as opposed to the trustworthiness of each view individually.
In the uncertainty typology detailed by MacEachren et al.~\cite{MacEachren2012Visual}, trust is decomposed into three high-level types: (i) \emph{accuracy}, defined as correctness or freedom from mistakes, conformity to truth or to a standard or model; (ii) \emph{precision}, defined as the exactness or degree of refinement with which a measurement is stated or an operation is performed; and (iii) \emph{trustworthiness}, defined as source dependability or the confidence the user has in the information. The latter is a broad category that includes components such as completeness, consistency, lineage, currency/timing, credibility, subjectivity, and interrelatedness.

\begin{figure*}[tb]
 \centering 
 \includegraphics[width=\textwidth]{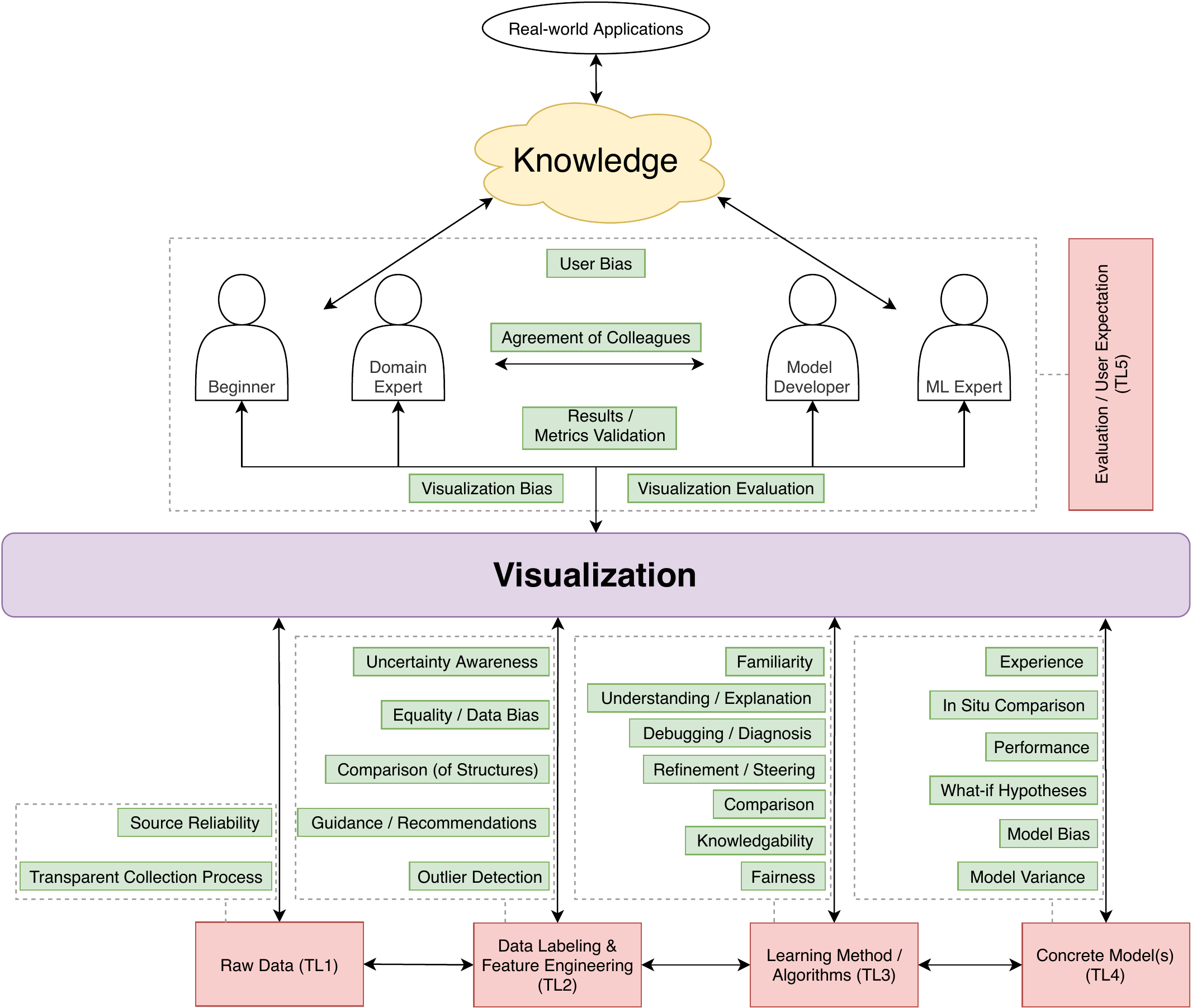}
 \caption{A typical ML pipeline (depicted in red), assisted by visualization (in purple). Issues of trust permeate the complete shown pipeline, and we locate and categorize these issues in several \textbf{trust levels (TLs)}. The various categories proposed in this work are represented in green. The yellow ``cloud'' represents the knowledge created by the different target groups while they pursue their goals by using visualizations to explore the pipeline, the data and/or the ML models. Finally, at the very top, we encode the real-world applications with an ellipsoid.}
 \label{fig:trust}
 \vspace{-.5em}
\end{figure*}

\noindent \textbf{Online questionnaire.} \quad 
The next step of our work can be compared to domain problem characterization~\cite{Munzner2009A}. In order to elicit the expectations and suggestions from ML practitioners with regard to our problem, 
we distributed an online questionnaire titled ``How Would Visualization Help Enhancing Trust in Machine Learning Models?'' (see supplementary material \textbf{S1}).
We received answers from 27 participants, all with at least a Bachelor's degree, and most with a Master's (40.7\%) or a Doctorate degree (51.9\%). Almost all of them had their education in Computer Science or related fields. Some participants have only used ML in a few projects (around 33.3\%), but most are either ML practitioners (22.2\%) or developers/researchers in the field (44.4\%). 
Their experiences with different types of ML algorithms/models are diverse, with rather balanced numbers between supervised (85.2\%) and unsupervised (70.4\%) learning. Within these two categories, classification (95.7\%) and clustering (89.5\%) are the most popular, respectively. 
The questionnaire itself begins with a description of a hypothetical scenario where a real-world data set was used (Pima Indians Diabetes, obtained from the UCI ML repository~\cite{Dua2017UCI}). Each of the 15 questions presents a possible use of visualization related to trust in ML, and participants are asked to score them from 1 (strong disagreement) to 5 (strong agreement).
Questions are also accompanied by short descriptions of some characteristics of the proposed scenario, in order to help participants in answering them.

According to the results, the bulk of the answers in most of the questions is concentrated around the scores of 4 and 5. This is evidence that the overall attitude of the participants towards visualization for enhancing trust in ML is largely positive.
Factors such as visualizing details of the source of the data (Q1), data quality issues (Q3), performance comparison of different ML algorithms (Q4), hyper-parameter tuning (Q5), exploration of ``what-if'' scenarios (Q11), and investigation of fairness (Q12) obtained the majority of votes on score 5.
Other factors which showed very positive---but less overwhelming---opinions were the visualization of details about the data collection process (Q2), data control and steering during the training process (Q6 and Q9), feature importance (Q7), visualizing the decisions of the model (Q8 and Q10), enabling collaboration (Q13), and the choice of tools for specific models (Q14). In these cases the majority of the scores were 4, but with some variance towards 3 and 5.
The only question that deviated from this trend was the last one (Q15), where we proposed that a single well-designed performance metric would be enough to judge the quality of an ML model, and no further actions (such as visualization) would be necessary. In this case, most of the scores were concentrated on either 1 or 2, showing clear disagreement.

\begin{table}[tb]
 \centering 
 \caption{Summary of the answers to the two open-ended questions on the participants' expectations and suggestions, which were provided at the end of the online questionnaire. The answers are sorted based on the number of occurrences \# and then alphabetically.}
 \includegraphics[width=.9\columnwidth]{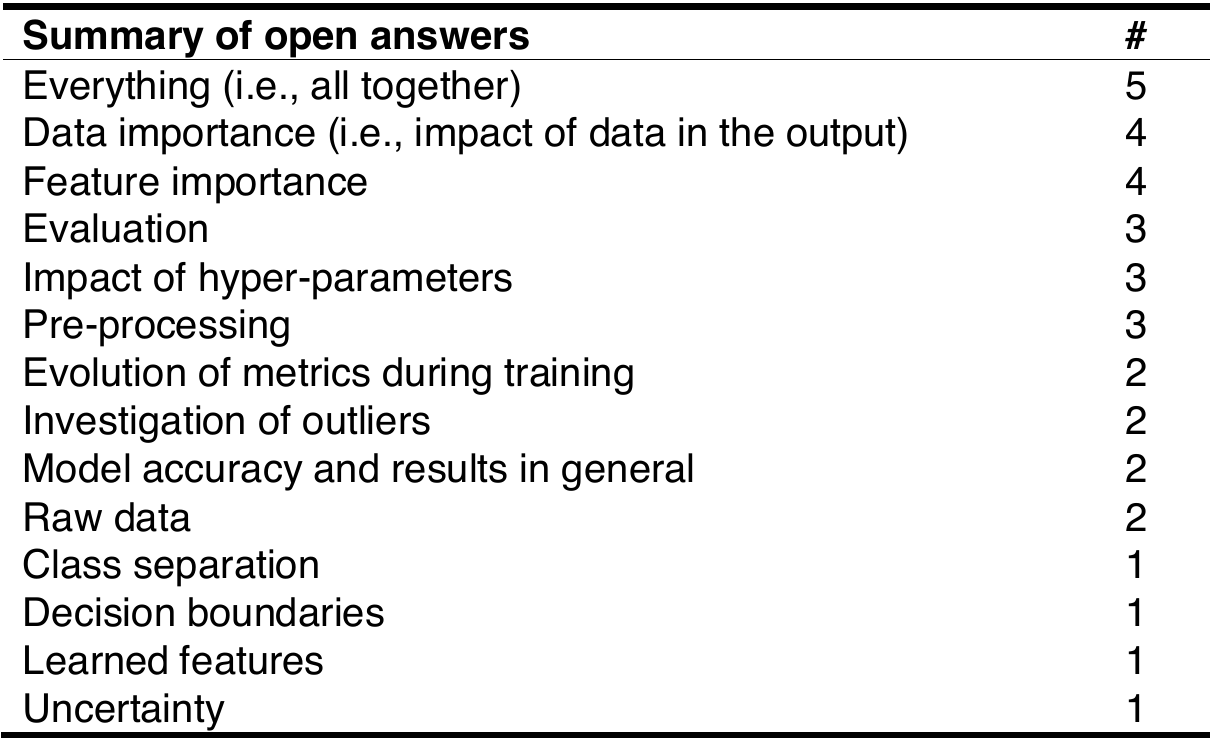}
 \label{tab:quest_ans}
 \vspace{-.5em}
\end{table}

The questionnaire ends with two open-ended questions, where participants were free to give their ideas and opinions on which steps of the ML process (or properties of the models and the data) they would like to visualize to increase the trust in the ML models they use. 
Many participants indicated their desire to visualize the ML process as much as possible, in all phases where it might apply (5 answers). Additionally, out of all the specific concepts and ideas that emerged, the most popular were the visualization of feature importance (4 answers), the impact of different characteristics of the data instances (4 answers), investigation of hyper-parameters (3 answers), visualizing the pre-processing steps (3 answers), and the evaluation of the model (3 answers). \autoref{tab:quest_ans} summarizes all these answers along with the number of occurrences. 
These answers were mostly aligned with our prior hypotheses, but also enabled us to gain new insights on what was missing from our categorization of trust factors (see below).
For instance, the \emph{source reliability} category was influenced by one participant who described her/his work with Parkinson's disease data and the reliability problems involved with it: ``For instance, I have been working with clinical studies with Parkinson's disease patients wearing sensors in their wrists. For us researchers, it was difficult to see how the data was collected e.g. patients could do a certain daily activity (e.g. cutting grass) but in our model we accounted that as tremor.'' 
Another important point that was brought up is that visualization-based steering of the ML training process might push the user to \cit{fish} for desired results and invalidate the statistical significance of the model. 

\noindent \textbf{Trust levels (TLs) and categories.} \quad 
In this STAR, we cover the subject of \emph{enhancing trust in ML models} with the use of
\emph{visualizations}. As such, we do not cover solutions proposed to address
those questions solely at the algorithmic level, even if they are considered
with growing interest by ML researchers (as exemplified by the two plenary
invited talks \cite{Howard2018Investigations,Spiegelhalter2018Making} on the subject given in 2018 at NeurIPS, one of the major ML venues). 
Based on the existing work discussing the issues of trust, the suggestions from ML experts (see above), and internal discussions, we consider that the problem of enhancing trust in ML models has a multi-level nature.
It can be divided into five \textbf{TLs} related to trustworthiness of the following: \emph{the raw data} ($\rightarrow$TL1), \emph{the processed data} ($\rightarrow$TL2), \emph{the learning method (i.e., the algorithms)} ($\rightarrow$TL3), \emph{the concrete model(s) for a particular task} ($\rightarrow$TL4), and \emph{the evaluation and the subjective users' expectations} ($\rightarrow$TL5). These levels of trust are aligned with the usual data analysis processes of a typical ML pipeline, such as (1) collecting the raw data; (2) allowing the user to label, pre-process, and query/filter the data; (3) interpreting, exploring, and explaining algorithms in a transparent fashion; (4) refining and steering concrete model(s); and (5) evaluating the results collaboratively. With the term \emph{algorithm}, we define an ML method (e.g., logistic regression or random forest); in contrast to a \emph{model} which is the result of an algorithm and is trained with specific parameters. 

We use the term \emph{level} to refer to the increasingly abstract nature of
concepts as well as to emphasize the sequential aspect of the ML
pipeline. Indeed, the lack of trustworthiness in each stage of the pipeline cumulatively introduces instability in the predictions of a model. Thus, trust issues (i.e., categories) that are relevant to two or more of our TLs are assigned to the lowest TL possible. This is similar to concerns about issues cascading from earlier to later levels within the nested model for visualization design and validation~\cite{Munzner2009A}, for instance. \autoref{fig:trust} displays the connection between a typical ML pipeline (slightly adapted from the work of Sun et al.~\cite{Sun2017Label}) and the visualization techniques that enhance trust in ML models in various phases. In the bottom layer of \autoref{fig:trust}, we depict (in red) the ML pipeline comprising the distinct areas where users are able to interact with (and choose from) a large pool of alternatives or combinations of options. The layer above depicts the visualization (in purple).
The upper layer consists of the different target groups that we address, the generation of knowledge, and the usability of this knowledge in solving problems stemming from real-life applications. 
Finally, the multiple categories of trust associated with each of these levels are presented in green in \autoref{fig:trust} and discussed in detail below.

\begin{itemize}
\vspace{.5em}
\item \textbf{Raw data (TL1).} \quad 
The lowest trust level gathers categories attached to the data collection itself. 
They belong to the complex task of preparing the data for further analysis, commonly referred to as \emph{data wrangling}~\cite{Kandel2011Research}.

Arguably, \emph{source reliability} is the very first category that should be visualized in a system. 
It should detect and handle the cases that do not meet the quality expectations or that show unusual behavior. 
For instance, detecting that some labels are unreliable could guide the user in selecting ML algorithms that are resistant to label noise \cite{Frenay2014Classification}.
However, perceiving source reliability is not an easy task, as it involves visualization questions, such as ``how to visualize the data source involved in data collection?'', but also the very statistical questions of measuring reliability.

As a proxy for this measure, one can visualize information, for instance, ``was a particular university involved in data collection, was a domain expert such as a doctor present during the health data collection, and were the sensors reliable and error-free?'' 
Hence, source reliability is strongly related to ensuring a \emph{transparent collection process}, the second category of this level. 
This includes visualizing the data collection process, what systems were used to collect the data, and how, why, and how objectively that was done.

Issues about reliability of the data and of the collection process can jeopardize, from the very start, the ML process and diminish the TLs set by users. 
If those issues remain undetected, they can spoil the later phases, according to the classic ``garbage in, garbage out'' principle. 
For instance in the case of unreliable labels \cite{Frenay2014Classification}, reported error rates are also unreliable. 
This is becoming more relevant with the growing attention given to adversarial machine learning, an ML research field which focuses on adversarial inputs designed to break ML algorithms or models~\cite{10.1145/3134599,Laskov2010}.

\vspace{.5em}
\item \textbf{Data labeling \& feature engineering (TL2).} \quad
The next group of categories has its focus beyond the raw data and into feature engineering and labeling of the data. 
This is also partially related to \emph{data wrangling}. 
Trust issues at this level focus on data that are overall considered to be reliable and clean. 
Trust can then be enhanced by addressing subgroup or instance problems.

With \emph{uncertainty awareness} and visualizations supporting it, the data instances that do not fit can be filtered out, and any borderline cases are highlighted to be explored by the users via visual representations.

The category \emph{equality/data bias} is related to the \emph{fairness} category discussed below. 
It concerns the possible sources of subgroup-specific bias in the decision of an ML model. 
For instance, if a subgroup of the population has characteristics that are significantly different from the ones of the population as a whole, then the decisions for   members of this subgroup could be unfair compared to the decisions for members of other subgroups. 
Visualization methods can be used to explore interesting subgroups and to pinpoint potential issues.

\emph{Comparison (of structures)}~\cite{Kim2017Comparison} implies the usage of visualization techniques in order to compare different structures in the
data. 
As an example, experts in the biology domain would like to compare different structures visually, and furthermore, improve these representations with various encodings such as color.

\emph{Guidance/recommendations}~\cite{Ceneda2019A} is a good continuation of the previous concept: trust can be improved by using visualization tools that (1) recommend new labels in the unlabeled data scenarios, for example, in semi-supervised learning and (2) guide the user to manage the data by adding, removing, and/or merging data features and instances.

Finally, for this level of trust, \emph{outlier detection}, i.e., searching and investigating extreme values that deviate from other observations of a data set, can be alleviated by visualization systems (this is a major issue in ML~\cite{Chandola2009Anomaly}). 
Detecting and manipulating in a meaningful way an observation that diverges from an overall pattern on a sample is a useful way to positively influence the results and boost overall trust in the process. 
Notice that this category focuses on particular instances, while the \emph{source reliability} category described previously, considers data globally.

\vspace{.5em}
\item \textbf{Learning method/algorithms (TL3).} \quad
This group of categories concerns the ML algorithms themselves, as the third step of the ML pipeline. 
Each category corresponds to a particular way of enabling a better control, in broad sense, over ML algorithms.

\emph{Familiarity} is how visualization can support users in order to help them getting familiar with a certain learning method. 
There is a possibility that users are biased towards using an ML algorithm they know instead of the others that might actually be more appropriate.
Improvement of familiarity by using visualization could help to both limit this type of bias and to enhance the users' trust in algorithms they do not know well.

Interpretability and explainability are among the most common and widespread categories---being found in most of the papers that we identified. We further subdivide both into the following categories:
\begin{itemize}
    \item understand the reasons behind ML models' behavior and why they deviate from each other (\emph{understanding/explanation});
    \item diagnose causes of unsuccessful convergence or failure to reach a satisfactory performance during the training phase (\emph{debugging/diagnosis});
    \item guide experts (and novices) to boost the performance, transparency, and quality of ML models (\emph{refinement/steering}); and
    \item compare different algorithms (\emph{comparison}).
\end{itemize}
It should be noted that the issue of interpretability and explainability has been receiving a growing attention in the ML community. Algorithms are modified in order to produce models that are easier to interpret. However, those models are frequently claimed to be more interpretable based on general rules of thumb, such as ``rule-based systems are easier to understand than purely numerical approaches'' or ``models using fewer features than others are easier to understand''. Only the most recent papers tend to include user-based studies~\cite{pmlr-v80-adel18a,pmlr-v80-chen18j}. Unfortunately, they only explore quite simple visualization techniques such as static scatterplots. 
    
\emph{Knowledgeability} translates to the question: if users are not aware of an ML algorithm, then how are they supposed to use it? 
Possible solutions to provide assistance to users in such situations include visualizations designed to compare different models or to provide details about each algorithm. 
However, the lack of visualization literacy limits the possibilities for exploration of an ML algorithm and effects negatively all the categories of this phase~\cite{Boy2014A,Borner2019Data}. 
Model-agnostic (more general) visualization techniques that consider multiple algorithms can also support this challenge.

Last but not least, the category of \emph{fairness} covers the analysis of subgroup-specific effects in ML prediction, e.g., whether predictions are equally accurate for various subgroups; for instance, females versus males, or if there are discrepancies that give a group an advantage or a disadvantage compared to other groups. This topic has recently received a lot of attention in the ML community. It has been shown in particular that the most natural fairness and performance criteria are generally incompatible~\cite{kleinberg_et_al:LIPIcs:2017:8156}. Thus, ML algorithms must make some compromises between those criteria which justify the strong need for visually monitoring/analyzing such trade-offs.

\vspace{.5em}
\item \textbf{Concrete model(s) (TL4).} \quad
This final step of the ML pipeline consists of turning its inputs, mainly a set of ML learning methods/algorithms, into a \emph{concrete model or a combination of models} \cite{Sacha2019VIS4ML}. 
Trust issues related to this step concern mostly performance related aspects, both in a static interpretation but also in a dynamic/interactive way.

\emph{Experience} is a primary crucial factor, since promoting personalized visualizations based on the experiences of a user alter and might determine the selection. 
As an example, an expert in ML, a novice user, and a specific domain expert have different needs, and ``what are their experiences and how can the visualization adapt to that?'' is an important question.

\emph{In situ comparison} can be described as comparing different snapshots and/or internal structures of the same concrete model in order to enhance trust.

\emph{Performance} is another very common way to monitor the results of a model visually. 
Performance can objectively compare a model with another. 
However, this is usually insufficient for a complete understanding of the trade-off between different models.

\emph{What-if hypotheses} appear when users search for impacts based on their interactions. 
A potential question is: ``What is the consequence if we change one parameter and keep the rest stable for a specific model, or select some points to explore further?''

\emph{Model bias} and \emph{model variance} are well-known concepts originating from statistics with regard to the bias-variance trade-off. The \emph{bias} is a systematic error created by the wrong hypotheses in a model. High bias can cause a model to avoid seeing the relevant associations between features and target outputs, thus underfitting. The \emph{variance} is a manifestation of the model's sensitivity or the lack thereof to the data, more precisely to the training set. It could also be the result of parameterizations or perturbations. High variance can result in a model which bears inside random noise in the training data, rather than the intended outputs, hence overfitting.

\vspace{.5em}
\item \textbf{Evaluation/user expectation (TL5).} \quad
The last group of categories is subsequent to visualization tools and techniques that oversee the ML pipeline, leading to knowledge generation in the overall workflow. 
Evaluation of models and meeting user expectations~\cite{Chuang2012Interpretation} is a key component for people to trust or not ML model(s) for a task.

\emph{Agreement of colleagues} is supported by visualizations with provenance~\cite{Oliveira2017A,Ragan2016Characterizing} and collaborative visualizations which facilitate, for instance, ten experts from diverse domains to agree that a model performed well. 
This purpose could be served by provenance features and specific glyphs or snapshots, along with web-based online tools and platforms. 
When using visualizations, the choices of the visual metaphor and the visual variable (e.g., color instead of size) are important but can supplement the process negatively with \emph{visualization bias}. This kind of bias was described, for example, by Lespinats and Aupetit~\cite{Lespinats2011CheckViz}. 
However, this issue is being addressed by multiple ongoing research efforts in various subfields of visualization which are outside of the scope of this survey~\cite{Mayr2019Trust,Xiong2019Examining}. 
Thus, we have not included this perspective in our categorization.

A measure against visualization bias that we consider instead is the \emph{visualization evaluation} \cite{Munzner2009A} that many authors of visualization papers perform. 
Quantitative or qualitative methods are used in the InfoVis and other communities to evaluate new visualization techniques. 
Both count as visualization evaluations, even receiving feedback from ML experts and/or domain experts before, during, or after the development of a visualization system. 

Moreover, \emph{results/metrics validation} is the most common method utilized by developers of visualization tools to indicate if a model can be trusted and has reached user expectations. 
However, we believe that it is not sufficient on its own.

Finally, \emph{user bias} is a rarely addressed category which tries to understand the cognitive biases of users who have the power to steer an automated process. Questions such as where, when, and why a user has to interact with a model are still an open challenge. A paper from Nalcaci et al.~\cite{Nalcaci2019Detection}, for example, works with distinction bias and confirmation bias in visualization systems that are related to user bias when viewing visualizations. Also, a recent survey from Dimara et al.~\cite{Dimara2020A} tries to connect the possibly-biased judgment and decision making of humans with specific visualization tasks.
\end{itemize}

\section{Related Surveys} \label{sec:relwo}
	The challenge of enhancing trust in ML models has not yet received the same level of attention in systematic surveys as other topics, for example, the understanding and interpretation of DR or deep learning (DL). 
To the best of our knowledge, this is the only survey that deals with InfoVis and VA techniques for enhancing the trustworthiness of ML models. 
In order to confirm that, we carefully examined the online browser of the survey of surveys (SoS) paper from McNabb and Laramee~\cite{McNabb2017Survey}, which contains 86 survey papers from main InfoVis venues. We have also investigated 18 additional survey papers in our own recent SoS paper~\cite{Chatzimparmpas2020A}. Our analysis indicated that many of these surveys are about interpretable ML models, especially regarding current popular subjects such as \emph{interpretable and interactive ML (IML)}, \emph{predictive VA (PVA)}, \emph{DL}, and \emph{clustering and DR}.
None of these papers, however, has an explicit focus on categorizing and/or analyzing techniques related especifically to the subject of trust in ML models. Related issues, such as accuracy, quality, errors, stress levels, or uncertainty in ML models, are touched upon by some of them, but in our work these issues are discussed in more detail. Particularly, uncertainty in the data and the visualization itself is a part of our TLs in the \emph{uncertainty awareness} and the \emph{visualization bias} categories. One of the main differences in our work is the focus on the transformation from uncertainty to trust, which should happen progressively in all phases of an ML pipeline. Some previous works offer brief literature reviews and propose frameworks for human-centered IML with visualization~\cite{Sacha2016Human,Sacha2017What}, the problem of integrating ML into VA~\cite{Endert2017The}, trust in VA~\cite{Sacha2016The}, or comparison of DR techniques from an ML point of view~\cite{vandermaaten2009Dimensionality}. Although interesting, those papers fall outside the scope of the \emph{trust in ML models} subject. One of the motivations for this STAR came from our analysis of the future work sections of these surveys---10 out of the 18 surveys highlight the subject of enhancing trust in the context of ML models, making this challenge one of the most emergent and essential to solve. This body of work also forms the basis for our methodological part of the literature research, presented in \autoref{sec:meth}.

\subsection{Interpretable and Interactive Machine Learning}

The work concerning the interpretability of ML models in the visualization community started to emerge around 15 years ago. 
This opportunity was captured by Liu et al.~\cite{LiuW2017Towards} who conducted a survey that summarizes several ML visualization tools focusing on three categories (\emph{understanding}, \emph{diagnosis}, and \emph{refinement}). This is different compared to our perspective and goal to categorize only those papers that tackle the problem of enhancing trust in ML models. 
The recent publication by Du et al.~\cite{Du2020Techniques} groups techniques for interpretable ML into \emph{intrinsic} and \emph{post-hoc}, which can be additionally divided into \emph{global} and \emph{local} interpretability. The authors also suggest that these two types of interpretability bring several advantages, for example, that users trust an algorithm and a model's prediction. 
However, they do not analyze in details the different aspects of enhancing trust in ML models as we performed in this STAR. 
Overall, these surveys (and the categories from Liu et al.~\cite{LiuW2017Towards}, together with \emph{comparison}) target the interpretability and explainability at the level of ML algorithms, which are themes under the umbrella of VIS4ML (visualization for ML) and comprise only a small subset of our proposed categorization.

Moreover, the topic of IML aided by visualizations has been discussed in many papers recently, as it was summarized in the surveys by Amershi et al~\cite{Amershi2014Power} and Dudley and Kristensson~\cite{Dudley2018A}. The former focused on the \emph{role of humans in IML} and \emph{how much users should interfere and interact with ML}. They also suggested at which stages this interaction could happen and categorized their papers accordingly. 
Steering, refining, and adjusting the model with domain knowledge are not trivial tasks and can introduce cumulative biases into the process. 
Due to this, in this STAR our analysis focuses on the biases that a user might introduce into a typical ML pipeline.  Furthermore, visualizations may introduce different biases to the entire process, as discussed in the previous \autoref{sec:back}. In such situations, the visualization design should be directed towards conveying, or occasionally removing, any of these biases initially and not simply making it easier for users to interact with ML models.

\subsection{Predictive Visual Analytics}

Lu et al.~\cite{Lu2017The} adopted the pipeline of PVA, which consists of four basic blocks: (i) \emph{data pre-processing}, (ii) \emph{feature selection and generation}, (iii) \emph{model training}, and (iv) \emph{model selection and validation}. These are complemented by two additional blocks that enable interaction with the pipeline: (v) \emph{visualization} and (vi) \emph{adjustment loop}. The authors also outline several examples of quantitative comparisons of techniques and methods before and after the use of PVA. 
However, no analysis has been performed about trust issues that are incrementally added in each step of the pipeline. Another survey written by Lu et al.~\cite{LuC2017Recent} follows a similar approach by classifying papers utilizing the same PVA pipeline, but with two new classes: (a) \emph{prediction} and (b) \emph{interaction}. For instance, regression, classification, clustering, and others are the primary subcategories of the prediction task; and explore, encode, connect, filter, and others, are subcategories of interaction. This work inspired us to introduce the \emph{interaction technique} subcategory of our basic category, called \emph{visualization}. One unique addition, though, is the \emph{verbalize} category, which describes how visualization and use of words can assist each other by making the visual representation more understandable to users and vice versa. Concluding, none of these survey papers so far provide future opportunities touching the subject of how visualization can boost ML models' trustworthiness.

\subsection{Deep Learning}

Gr{\"u}n et al.~\cite{Grun2016A} briefly explain how the papers they collected are separated to their taxonomy for feature visualization methods. The authors defined three discrete categories as follows: (i) \emph{input modification methods}, (ii) \emph{deconvolutional methods}, and (iii) \emph{input reconstruction methods}. Undoubtedly, learned features of convolutional neural networks (CNNs) are a first step to provide trust to users for the models. But still, this step belongs to the interpretability and explainability of a specific algorithm, i.e., very specialized and targeted to CNNs. In our work, we cover not only CNNs but every ML model with a focus on the data, learning algorithms, concrete models, users, and thus not only on the model. The two main contributions of Seifert et al.~\cite{Seifert2017Visualizations} are the analysis of insights that can be retrieved from deep neural network (DNN) models with the use of visualizations and the discussion about the visualization techniques that are appropriate for each type of insight. In their paper, they surveyed visualization papers and distributed them into five categories: (1) \emph{the visualization goals}, (2) \emph{the visualization methods}, (3) \emph{the computer vision tasks}, (4) \emph{the network architecture types}, and (5) \emph{the data sets that are used}. This paper is the only one that contains analyses for the data sets used in each visualization tool, which worked as a motivation for us to include a data set analysis in our survey. However, their main contributions do not touch the problem of trustworthiness, but more the correlation of visualizations and patterns extraction (or insights gaining) for DNNs. A summarization of the field of interpreting DL models was performed by Samek et al.~\cite{Samek2017Explainable}, putting into the center the increasing awareness of how interpretable and explainable ML models should be in real life. 
The main goal of their survey is to foster awareness of how useful it is to have interpretable and explainable ML models.  
General interpretability and explainability play a role in increasing trustworthiness, but not a major one. The different stages of the ML pipeline should be taken into account as from early stages, bias and deviance can occur and grow when processing through the pipeline. Zhang and Zhou~\cite{Zhang2018Visual} study their papers starting from the visualization of CNN representations between network layers, over the diagnosis of CNN representations, and finally examining issues of disentanglement of ``the mixture of patterns'' of CNNs. 
They neither provide a distinct methodology of categorization for their survey, nor insights on the problem of trust as opposed to our survey.

Another batch of papers on DL assembles into Garcia et al.'s~\cite{Garcia2018A} survey in which visualization tools addressing the interpretability of models and explainability of features are described. The authors focus on various types of neural networks (NNs), such as CNNs and recurrent neural networks (RNNs), by incorporating a mathematical viewing angle for explanations. They emphasize the value of VA for the better understanding of NNs and classify their papers into three categories: (a) \emph{network architecture understanding}, (b) \emph{visualization to support training analysis}, and (c) \emph{feature understanding}. 
In a similar sense, (i) \emph{model understanding}, (ii) \emph{debugging}, and (iii) \emph{refinement/steering} are three directions that Choo and Liu~\cite{Choo2018Visual} consider. Model understanding aims to communicate the rationale behind model predictions and spreads light to the internal operations of DL models. In cases when the DL models underperform or are unable to converge, debugging is applied to resolve such issues. Finally, model refinement/steering refers to methods that enable the interactive involvement of  usually experienced experts who build and improve DL models. 
Compared to our survey, only half of the learning methods are considered. Thus, their reader support is limited when it comes to show how their algorithms actually work on several occasions. Yu and Shi~\cite{Yu2018A} examined visualization tools that support the user to accomplish four high-level goals: (1) \emph{teaching concepts}, (2) \emph{assessment of the architecture}, (3) \emph{debugging and improving models}, and (4) \emph{visual exploration of CNNS, RNNs, and other models}. They describe four different groups of people in their paper: (a) \emph{beginners}, (b) \emph{practitioners}, (c) \emph{developers}, and (d) \emph{experts}, distributed accordingly to the four aforementioned classes. These groups are also considered in our work. Nonetheless, teaching concepts and assessing the architectures of DNNs are particular concepts that do not enhance trust explicitly. This is why we focus on multiple other categories, such as models' \emph{trade-off of bias and variance} or \emph{in situ comparisons of structures of the model}, in general and not exclusively for DL models. Hohman et al.~\cite{Hohman2019Visual} surveyed VA tools that explore DL models by investigating papers into six categories answering the aspects of ``who'', ``why'', ``what'', ``when'', ``where'', and ``how'' of the collected papers. Their main focus is on interpretability, explainability, and debugging models. 
The authors conclude that just a few tools visualize the training process, but solely consider the ML results. Our \emph{ML processing phase} category is motivated by this gap in the literature, i.e.,
we investigate this challenge in our paper to gain new insights about the correlation of trust and visualization in \emph{pre-processing}, \emph{in-processing}, and \emph{post-processing} of the overall ML processing phases. Finally, as many explainable DL visualization tools incorporate clustering and DR techniques to visualize DL internals, the results of these methods should be validated on how trustworthy they are.

\subsection{Clustering and Dimensionality Reduction}

Sacha et al.~\cite{Sacha2017Visual} propose, in their survey, a detailed categorization with seven guiding scenarios for interactive DR: (i) \emph{data selection and emphasis}, (ii) \emph{annotation and labeling}, (iii) \emph{data manipulation}, (iv) \emph{feature selection and emphasis}, (v) \emph{DR parameter tuning}, (vi) \emph{defining constraints}, and (vii) \emph{DR type selection}.
During the \emph{annotation and labeling} phase, for example, hierarchical clustering could assist in defining constraints which are then usable by DR algorithms. 
Nonato and Aupetit~\cite{Nonato2019Multidimensional} separate the visualization tools for DR according to the categories \emph{linear and nonlinear}, \emph{single- versus multi-level}, \emph{steerability}, \emph{stability}, and \emph{others}. Due to the complexity of our own categorization and our unique goals, we chose to use only their first category (\emph{linear versus nonlinear}), as is common in previous work~\cite{vandermaaten2009Dimensionality}. Nonato and Aupetit also describe different quality metrics that can be used to ensure trust in the results of DR. However, as the results of our online questionnaire suggested (cf.~\autoref{sec:back}), comparing those quality metrics alone is probably not sufficient. To conclude, the main goal of these two surveys is not related to ML in general, and the latter one only discusses trust in terms of aggregated quality metrics. This is a very restricted approach when compared to our concept of trust, which should be ensured at various levels, such as \emph{data}, \emph{learning method}, \emph{concrete model(s)}, \emph{visualizations themselves}, and \emph{covering users expectations}.

\section{Methodology of the Literature Search} \label{sec:meth}
	In the following, we present the methodology used to identify and systematically structure the papers of our STAR. Our work is inspired by the same methodology guidelines from Lu et al.~\cite{Lu2017The}, Garcia et al.~\cite{Garcia2018A}, and Sacha et al.~\cite{Sacha2017Visual} presented in~\autoref{sec:relwo}. 
In an initial pilot phase (cf.~\cite{Snyder2019Literature}), we extracted appropriate keywords from ten relevant papers~\cite{Viechtbauer2015A,Whitehead2016Estimating}, including those that deal with the problems of interpretable/explainable ML (which are closely related to trust in ML).
The keywords were divided into two lists with the goal to cover both \emph{trust} and \emph{ML}.
For \emph{trust}, the used keywords were, in alphabetical order:  ``accuracy'', ``assess'', ``bias'', ``black box'', ``confidence'', ``diagnose'', ``distort'', ``error'', ``explain'', ``explore'', ``feedback'', ``guide'', ``interact'', ``noise'', ``quality'', ``robustness'', ``stress'', ``trust'' ``uncertainty'', ``validate'', ``verify'', and their derivatives. 
For \emph{ML}, the searched keywords were: ``artificial intelligence'', ``classification'', ``clustering'', ``deep learning'', ``dimensionality reduction'', ``machine learning'', ``neural network'', ``projections'', and all the types of ML (e.g., ``supervised learning'').

The keywords from the two lists were combined into pairs, such that each keyword from the first list was paired with each keyword from the second. These paired keywords were used for seeking papers relevant to the focus of this survey in different venues (cf. Section~\ref{sec:venues}).
A validation process was used in order to scan for new papers and admit questionable cases, as described in Section~\ref{sec:validation}. Papers that were borderline cases and eventually excluded are discussed in Section~\ref{sec:exclude}.

\subsection{Search and Repeatability} \label{sec:venues} 

To gather our collection of papers, we manually searched for papers published in the last 12 years (from January 2008 until January 2020). 
We started our search from InfoVis journals, conferences, and workshops, and later extended it to well-known ML venues (the complete list can be found at the end of this subsection).
Moreover, when seeking papers in ML-related venues (e.g., the International Conference on Machine Learning, ICML), we included two additional keywords: ``visual'' and ``visualization''. 

Within the visualization domain, we checked the following resources for publications:
\begin{description}
\item[Journals:] IEEE TVCG, Computers \& Graphics (C\&G), Computer Graphics Forum (CGF), IEEE Computer Graphics \& Applications (CG\&A), Information Visualization (IV), DiStill, and Visual Informatics (VisInf).

\item[Conferences:] IEEE Visual Analytics in Science and Technology (VAST), IEEE Visualization Conference (VIS) short papers track, Eurographics Visualization (EuroVis), IEEE Pacific Visualization (PacificVis), ACM Conference on Human Factors in Computing Systems (CHI), and ACM Intelligent User Interfaces (IUI).

\item[Workshops:] Visualization for AI Explainability (VISxAI), EuroVis Workshop on Trustworthy Visualization (TrustVis), International EuroVis Workshop on Visual Analytics (EuroVA), Machine Learning Methods in Visualisation for Big Data (MLVis), Visualization for Predictive Analytics (VPA), Visual Analytics for Deep Learning (VADL), IEEE Large Scale Data Analysis and Visualization (LDAV), and Visualization in Data Science (VDS).
\end{description}

\noindent Within the ML domain, we checked the following venues: 
\begin{description}
\item[Conferences:] ICML, Knowledge Discovery and Data Mining (KDD), and European Symposium on Artificial Neural Networks, Computational Intelligence, and Machine Learning (ESANN).
\item[Workshops:] ICML Workshop on Visualization for Deep Learning (DL), ICML Workshop on Human Interpretability in ML (WHI), KDD Workshop on Interactive Data Exploration \& Analytics (IDEA), NIPS Workshop on Interpreting, Explaining and Visualizing Deep Learning.
\end{description}

\noindent The search was performed in online libraries, such as IEEE Xplore, ACM Digital Library, and Eurographics Digital Library. As an example of the number of results we got, both IEEE Transactions on Visualization and Computer Graphics (TVCG) and IEEE Visual Analytics in Science and Technology (VAST) together resulted in around 750 publications. Due to the use of a couple of broad keyword combinations in order to cover our main subject effectively, some of the papers collected were not very relevant. They were sorted out in the next phase of our methodology.

\subsection{Validation} \label{sec:validation}

For the sake of completeness, we quickly browsed through each individual paper's related work section and tried to identify more relevant papers (a process known as \emph{snowballing}~\cite{Wohlin2014Guidelines}). With this procedure, we found more papers belonging to other venues, such as the Neurocomputing Journal, IEEE Transactions on Big Data, ACM Transactions on Intelligent Systems and Technology (ACM TIST), the European Conference on Computer Vision (ECCV), Computational Visual Media (CVM), and the Workshop on Human-In-the-Loop Data Analytics (HILDA), co-located with the ACM SIGMOD/PODS conference.
In more detail, this validation phase was performed in four steps:
\begin{enumerate}
\item we removed unrelated papers by reading the titles, abstracts, and investigating the visualizations; 
\item we split the papers into two categories: \emph{approved} and \emph{uncertain}; 
\item uncertain papers were reviewed by at least two authors, and if the reviewers agreed, they were moved to the approved papers;
\item for the remaining papers (i.e., where the two reviewers disagreed), a third reviewer stepped in and decided if the paper should be moved to the approved category or discarded permanently. 
\end{enumerate}
\noindent The calculated amount of disagreement, i.e., the number of conflicts in the 70 uncertain cases, was less than 20\% (approximately 1 out of 5 papers). This process led to 200 papers that made it into the survey.

\subsection{Borderline Cases} \label{sec:exclude}

We have restricted our search to papers with visualization techniques that explicitly focus on supporting \emph{trust} in ML models, and not on related perspectives (e.g., assisting the exploration and labeling process of input data with visual means). 
Therefore, papers such as those by Bernard et al.~\cite{Bernard2018Comparing,Bernard2018VIAL}, Gang et al.~\cite{Garg2010A}, and Kucher et al.~\cite{Kucher2017ALVA}, although undoubtedly interesting, are out of the scope of our survey, since their research contributions are exclusively based  on labeling data. 
Other partially-related papers~\cite{Abbas2019ClustMe,Ahmed2012An,Beham2014Cupid,Ferdosi2010Finding,Schreck2008Visual,Zhao2018BiDots} are also not included because they focus on using clustering solely to explore the data, without addressing inherent problems of the method. 
For similar reasons, the paper by Wenskovitch et al.~\cite{Wenskovitch2018Towards}, that tries to connect and aggregate benefits from clustering and DR methods, was excluded. 
Moreover, papers on high-dimensional data clustering or exploratory data analysis are not included (e.g., Behrisch et al.~\cite{Behrisch2014Feedback}, Lehmann et al.~\cite{Lehmann2015Visualnostics}, Nam et al.~\cite{Nam2007ClusterSculptor}, and Wu et al.~\cite{Wu2015EasyXplorer}). 
Finally, there are related works that provide important contributions to the visualization community, but do not study trust explicitly, and thus were not included: improving the computational performance of algorithms (e.g.,  t-SNE)~\cite{Pezzotti2016Hierarchical,Pezzotti2017Approximated}, frameworks and conceptual designs for closing the loop~\cite{Smith2018Closing}, investigating cognitive biases with respect to users~\cite{Choi2019Concept}, and enabling collaboration with the use of annotations~\cite{Chen2010Click}.

\section{General Overview of the Relations Between the Papers} \label{sec:overview}
	This section begins with a meta-analysis of the spatiotemporal aspects of our collection of papers. The analysis shows, on the one hand, that there is an increasing trend in trust-related subjects; on the other hand, it also highlights the struggles of collaborations between visualization researchers and ML experts. Additionally, we generated a co-authorship network to observe the connections of the authors from all the papers. By exploring the network and its missing links, we hope to bring researchers closer to form new collaborations towards research in the trustworthiness of ML models.

\begin{figure}[tb]
 \centering 
 \includegraphics[width=\columnwidth]{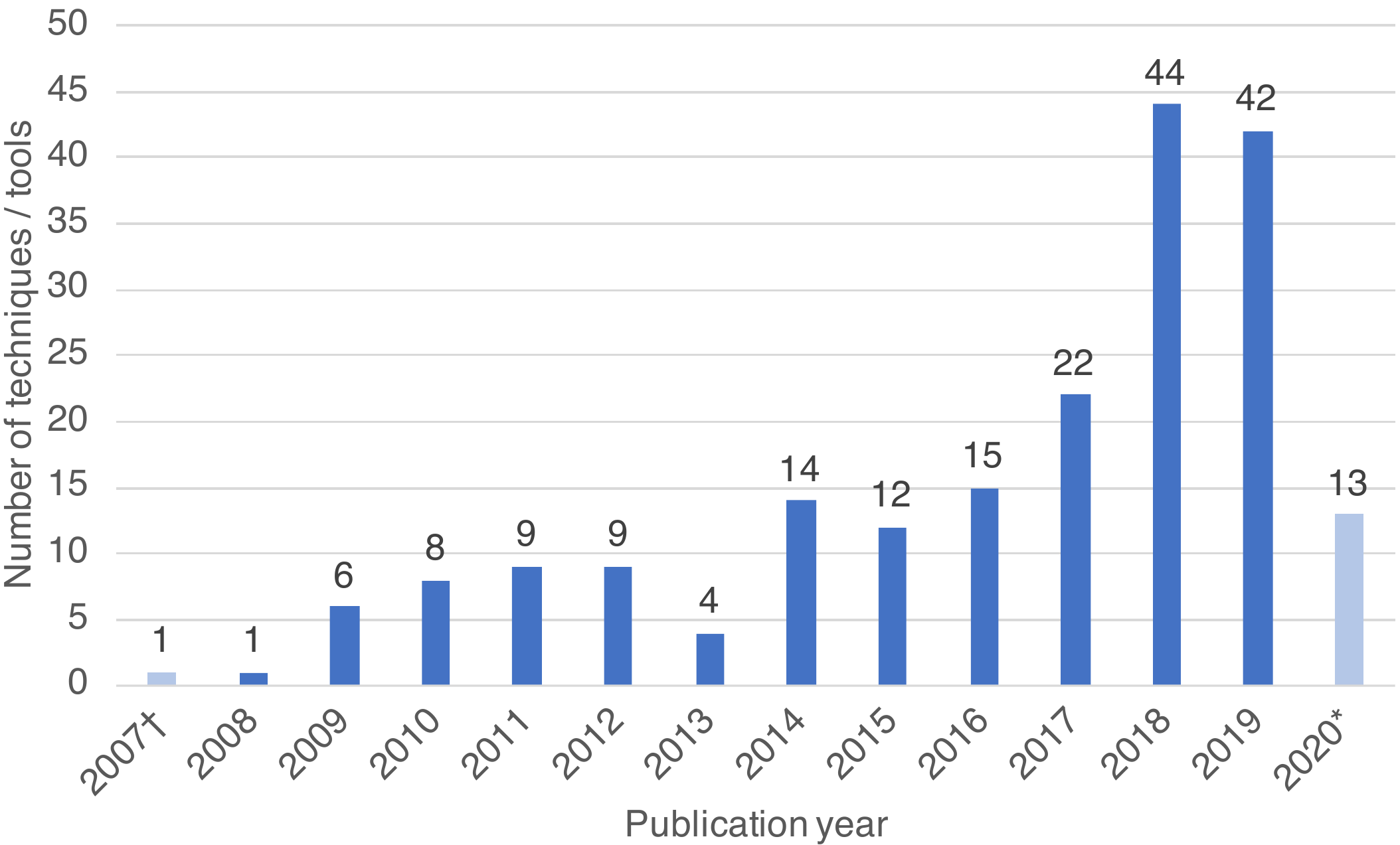}
 \caption{Histogram of the set of collected techniques/tools (200 in total) with regard to the publication year. ($\ast$) Please note that the data for 2020 is incomplete since the data collection for this survey was completed in January 2020. ($\dagger$) For 2007, we did not perform a complete search; the single publication was found within the related work section of another already-included paper.}
 \label{fig:time}
 \vspace{-.5em}
\end{figure}

\noindent \textbf{Time and venues.} \quad
Our collection of papers comprises 200 entries from a broad range of journals, conferences, and workshops. The analysis of the temporal distribution (see~\autoref{fig:time}) shows a stable growth in interest in the topic since 2009, with a sharp increase in 2018 and 2019 (and promising numbers also for 2020). 
The numbers for the publication venues identified can be seen in~\autoref{tab:venue}. 
Visualization researchers seem to be very interested in working with solutions to this problem and try to extend their work in ML venues with the creation of new workshops.
There is a large number of workshops on the topic, co-located with ML venues, which indicates that researchers are interested in reaching out of their respective areas in order to collaborate.  
However, the small number of publications outside of visualization venues could possibly show a struggle of visualization researchers to find and collaborate with ML experts. 
It might also indicate that ML experts are not fully aware of the possibilities that the visualization field provides.

\begin{table}[tb]
 \centering 
 \caption{Number of visualization techniques \# with regard to the respective publication venues in visualization (left and middle columns) and other disciplines (mostly ML venues; right column). Journals are marked with \lq{J}\rq and workshops with \lq{W}\rq. The remaining venues are conferences.}
 \includegraphics[width=\columnwidth]{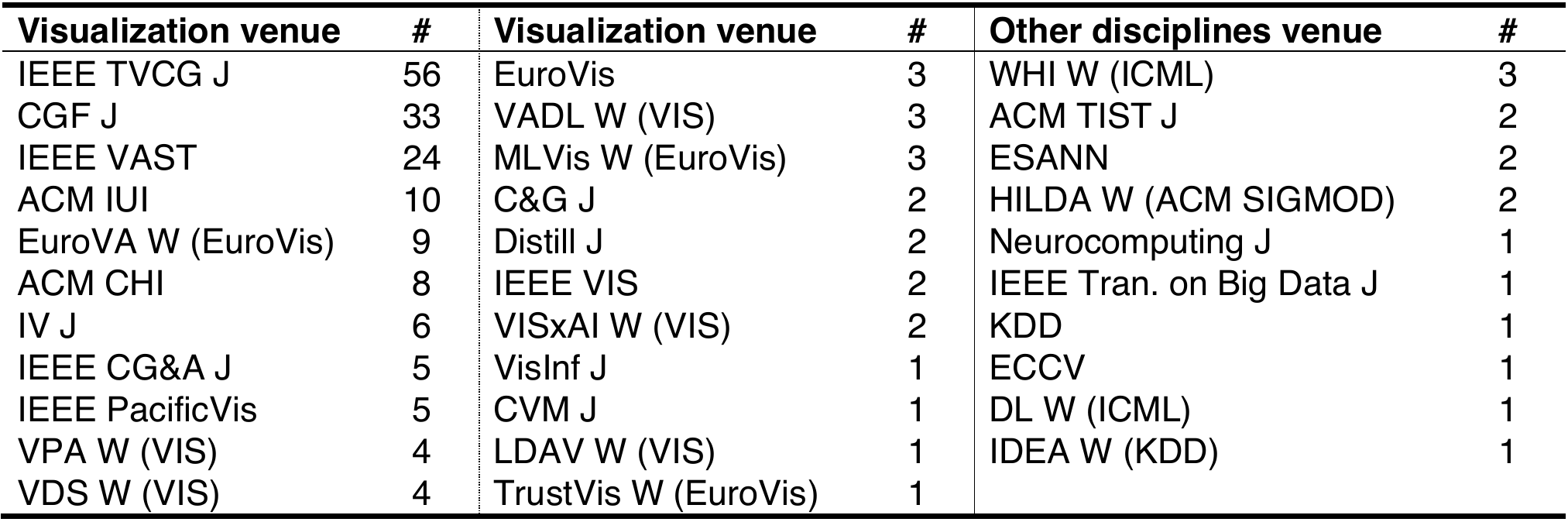}
 \label{tab:venue}
 \vspace{-.5em}
\end{table}

\begin{figure*}[htb!]
 \centering 
 \includegraphics[width=\textwidth]{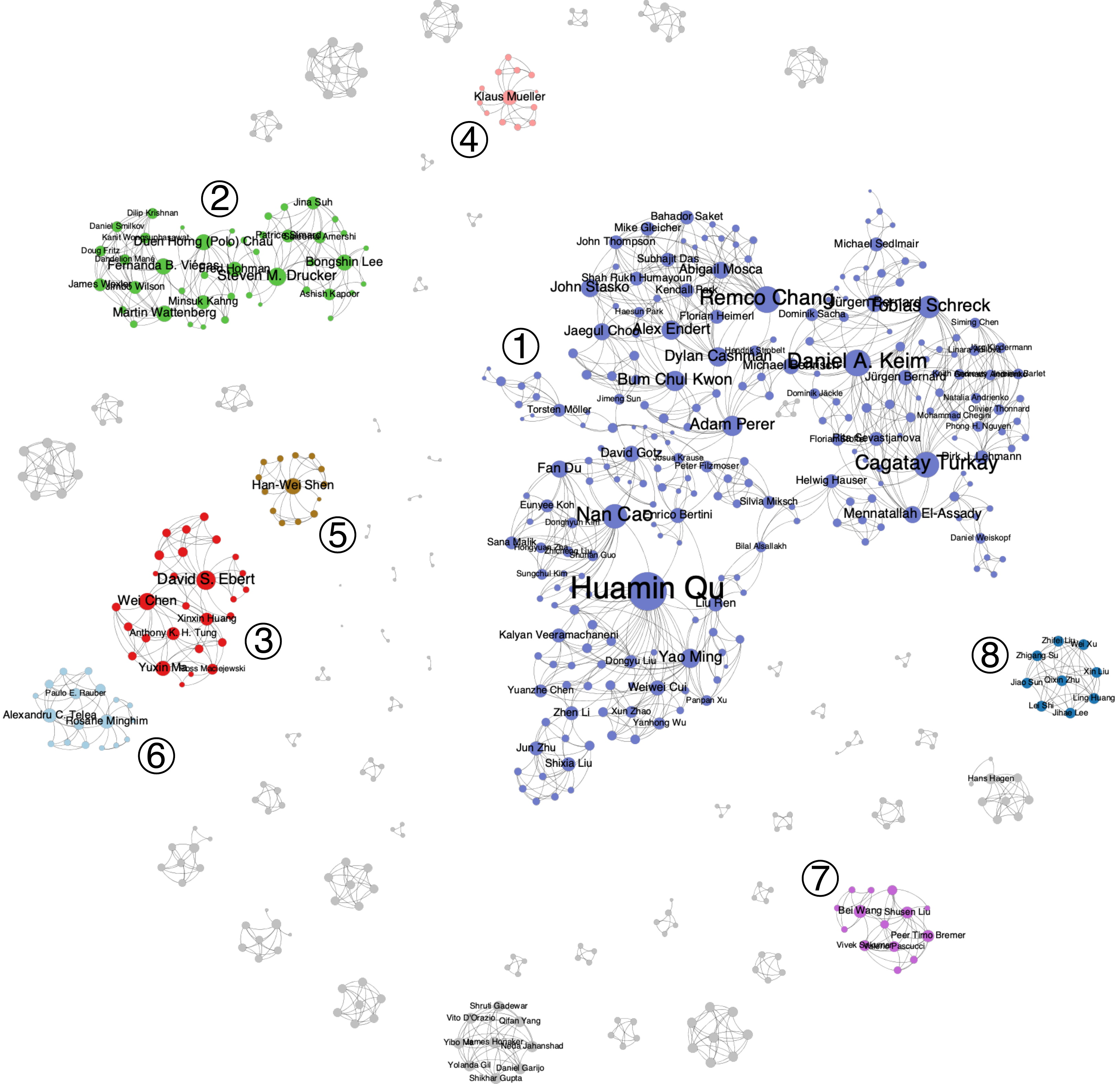}
 \caption{Co-authorship network visualization with the eight largest connected components (\circled{1}--\circled{8}) highlighted in different colors. 
 The node size represents the in-degree centrality of each author. The labels are filtered based on the in-degree value in order to reduce clutter.
 }
 \label{fig:coauthor}
 %
 %
\end{figure*}

\noindent \textbf{Co-authorship analysis.} \quad
We analyzed the co-authorship network of the authors of our collection of papers using Gephi~\cite{Bastian2009Gephi}, as presented in \autoref{fig:coauthor}. The goal was to identify a potential lack of collaboration within the visualization and ML communities. Enhancing collaboration between specific groups may lead to improvements in the subject of boosting trust in ML models with visualizations. The more connections an author has, the bigger is the size of the resulting node, i.e., the in-degree values of the graph nodes are represented by node size in the drawing. We colored the top eight clusters with the highest overall in-degree for all the nodes of each cluster. Finally, we filtered the node labels (authors first names and surnames) by setting a limit to the in-degree value in order to reduce clutter. By looking at the resulting co-authorship network (see \autoref{fig:coauthor} and \textbf{S2}), we can observe a huge cluster in violet \circled{1}. In this cluster, Huamin Qu, Remco Chang, Daniel A. Keim, Cagatay Turkay, and Nan Cao seem to be the most prominent authors, with many connections. If we consider different subclusters in this massive cluster, Nan Cao is the bridge between some of the subclusters. Another cluster on the left (with light green color \circled{2}) is related to the big industries (such as Google and Microsoft) with Fernanda B. Vi\'egas, Martin Wattenberg, and Steven M. Drucker as the most eye-catching names. Interestingly, this industry cluster is very well separated from the remaining academic clusters. 
Potentially, the connection of this industry cluster with the remaining clusters could have an impact on the research output produced by the visualization community.
There are many smaller clusters of collaborating people, for example, the cluster with David S. Ebert and Wei Chen \circled{3}, Klaus Mueller \circled{4}, Han-Wei Shen \circled{5}, Alexandru C. Telea \circled{6}, Valerio Pascucci \circled{7}, and others (e.g., \circled{8}) obviously serving as main coordinators.

\section{In-Depth Categorization of Trust Against Facets of Interactive Machine Learning} \label{sec:categ}
	In this section, we discuss the process and results of our categorization efforts. We introduce a multifaceted categorization system with the aim to provide insights to the reader about various aspects of the data and ML algorithms used in the underlying literature. The main sources of input for the categorization were the previous work from the surveys discussed in~\autoref{sec:relwo}, the iterative process of selecting papers (and excluding the borderline cases) described in \autoref{sec:meth}, and the feedback received from the online questionnaire (\autoref{sec:back}).
The top two levels of the proposed hierarchy of categories can be seen below, with 8 overarching aspects (6.1 to 6.8), partitioned into 18 category groups (6.1.1 to 6.6.3, plus TL1 to TL5), resulting in a total of 119 individual categories.

\begin{itemize}
\item 6.1. Data
\begin{itemize} \vspace{-1.0ex}
\item 6.1.1. Domain (10 categories)
\item 6.1.2. Target Variable (5 categories)
\end{itemize} \vspace{-1.0ex}
\item 6.2. Machine Learning
\begin{itemize} \vspace{-1.0ex}
\item 6.2.1. ML Methods (16 categories)
\item 6.2.2. ML Types (10 categories)
\end{itemize} \vspace{-1.0ex}
\item 6.3. ML Processing Phase (3 categories)
\item 6.4. Treatment Method (2 categories)
\item 6.5. Visualization
\begin{itemize} \vspace{-1.0ex}
\item 6.5.1. Dimensionality (2 categories)
\item 6.5.2. Visual Aspects (2 categories)
\item 6.5.3. Visual Granularity (2 categories)
\item 6.5.4. Visual Representation (19 categories)
\item 6.5.5. Interaction Technique (9 categories)
\item 6.5.6. Visual Variable (6 categories)
\end{itemize} \vspace{-1.0ex}
\item 6.6. Evaluation
\begin{itemize} \vspace{-1.0ex}
\item 6.6.1. User Study (2 categories)
\item 6.6.2. User Feedback (2 categories)
\item 6.6.3. Not Evaluated (1 category)
\end{itemize} \vspace{-1.0ex}
\item 6.7. Trust Levels (TL) 1--5
\begin{itemize} \vspace{-1.0ex}
\item TL1: Raw Data (2 categories)
\item TL2: Processed Data (5 categories)
\item TL3: Learning Method (7 categories)
\item TL4: Concrete Model (6 categories)
\item TL5: Evaluation/User Expectation (4 categories)
\end{itemize} \vspace{-1.0ex}
\item 6.8. Target Group (4 categories)
\end{itemize}
\noindent The complete overview of all categories is shown in \autoref{tab:categorizationLegend} (also in \textbf{S3} as a mind map). 
The aspect and category group names are preceded by the subsection numbers where they are introduced and discussed. This is to avoid confusion and to reduce the cognitive load of the reader.

\noindent \textbf{Designing the categorization.} \quad
Compared to previous surveys, we created new categories to better cover the 200 papers we included in our STAR. In the following list, we present the basic purpose for each aspect, along with the core similarities and differences when compared to the related surveys from \autoref{sec:relwo}.

\begin{itemize}
\item \textbf{6.1. Data} tries to create a link between the input data/application and the enhancement of trust in ML models.  
The first category group we identified in this aspect is the data \emph{domain}. We took the inspiration from our previous publication~\cite{Kucher2018The}, but the categories are significantly different to fit the current subject. 
In the case of the \emph{target variable}, we conceived the idea of separating the independent variable of each data set.
\item \textbf{6.2. Machine Learning} is an inherent component in boosting the trustworthiness of ML models. We used several sources for defining parts of the \emph{ML methods} category group. Different neural networks methods, such as CNNs, RNNs, DCNs, and  DNNs, were contained in other works~\cite{Seifert2017Visualizations,Yu2018A}. Also, as mentioned in \autoref{sec:relwo}, linear and non-linear DR is an existing categorization from \cite{Nonato2019Multidimensional}, but ensemble learning and the remaining DL categories are new to our STAR. \emph{ML types}, such as classification, regression, and clustering, can be seen in the work~\cite{Lu2017The}, but we improved this short categorization by using the complete supervised, unsupervised, semi-supervised, and reinforcement learning distinction.
\item \textbf{6.3. ML Processing Phase} connects the ML and the visualization aspects and shows when VA techniques are deployed to improve the trustworthiness of the ML models. \emph{During} and \emph{after training} categories can be found in the work of Hohman et al.~\cite{Hohman2019Visual}, which in our case are adjusted to the newly introduced \emph{pre-processing}, \emph{in-processing}, and \emph{post-processing} phases.
\item \textbf{6.4. Treatment Method} deals with differences between \emph{model-agnostic} or \emph{model-specific} approaches. Observing such distinctions might indicate where the community should later focus on to better boost the trust in ML models. \emph{Model-agnostic} $vs.$ \emph{model-specific} methods used in VA systems are first described in our work, although Dudley and Kristensson~\cite{Dudley2018A} hinted about \emph{model agnosticism}.
\item \textbf{6.5. Visualization} is another inherent component of how increasing trust in ML models can be achieved. Visualization details, such as \emph{dimensionality}, can also be found in the work of Kucher et al.~\cite{Kucher2015Text}. However, we added \emph{visual aspects and granularity}. \emph{Visual representation} was also inspired by Kucher et al.~\cite{Kucher2015Text} and many of the other related surveys. The \emph{verbalize} category is a novel addition to pre-existing work, which is part of the \emph{6.5.5. Interaction Technique} group described in the work of Lu et al.~\cite{LuC2017Recent}. Finally, for this aspect, the work of Kucher et al.~\cite{Kucher2018The} covers all the visual variables used by us except for \emph{opacity}.
\item \textbf{6.6. Evaluation} of visualization can reduce the visualization bias, thus boosting even further the application of VA systems for ML. We are among the first who included this new aspect to highlight the importance of evaluations in visualization systems, tools, and techniques.
\item \textbf{6.7. Trust Levels (TL) 1--5}, introduced in~\autoref{sec:back}, is the most novel category group. Only TL3, which contains the interpretability/explainability group of categories, is described in previous works~\cite{LiuW2017Towards,Choo2018Visual}. Despite that, \emph{comparison} is a fresh addition to this group.
\item \textbf{6.8. Target Group} is equally important to the problem of enhancement of the trustworthiness in ML models as the input (i.e., the data) and the actual visualization. This aspect is inspired mainly by Yu and Shi's paper~\cite{Yu2018A}.
\end{itemize}
\noindent Overall, this extensive categorization aims to completely unveil the relationship between trust and the remaining categories, as can be seen later in~\autoref{fig:correlation} and \autoref{sec:topic}.

\noindent \textbf{Filling in the categorization.} \quad 
To ensure consistency during the process of assigning the 200 papers to our categories in a first cycle, we created a code of practice (see \textbf{S4}) as a base structure. This base structure provides guidance to evaluate the individual papers in the same way without misalignment between the authors of this STAR. We also cleared and double-checked the resulting data for any issues that could come up with the annotated data in a second cycle. In particular, we looked for the following issues: (1) outliers, (2) typos, (3) discrepancies, and (4) inconsistencies between different evaluators by inspecting and removing any obscure and misclassified data cases. The fact that a large subset of the papers (75\%) were classified by the same author also maximizes the consistency of their final categorization. Each classified paper belongs to zero, one or more categories for every aspect depending on the information the paper contains. Due to the page limits and readability concerns, we cannot discuss all 200 papers in this section. Instead, we only focus on the most prominent and (in our opinion) most important ones. All 200 papers are referenced in Table~\ref{tab:topics} and are part of the bibliography. The complete survey data set, including the individual categorization for each paper, is provided in our online survey browser (see Fig.~\ref{fig:web}) and in \textbf{S5}.
%
\nocite{Jankowska2012Relative,Karer2018Panning,Krause2018AUser,Kulesza2015Principles,Micallef2017Interactive,Migut2010Visual,Rieck2016Exploring,Shao2017Interactive,Stolper2014Progressive,Strezoski2017Plug}

\begin{table*}[t]
\centering
\caption{The categorization used in our survey. The total number of corresponding visualization techniques per category is shown in every row, along with heatmap-style icons.}\vspace{-0.5em}
\setlength{\tabcolsep}{2pt}
\renewcommand{\arraystretch}{1}
\setlength\doublerulesep{2mm} 
\footnotesize
\begin{minipage}[t]{0.35\textwidth}
\begin{tabular}[t]{|l|r|}
\hline \textbf{6.1.1. Domain} & \textbf{198}~\statbox{0.01,0.51,0.01}{99\%}\\ 
\hline Biology & 28~\statbox{0.86,0.93,0.86}{14\%}\\ 
\hline Business & 19~\statbox{0.91,0.95,0.91}{10\%}\\
\hline Computer Vision & 59~\statbox{0.71,0.85,0.71}{30\%}\\
\hline Computers & 6~\statbox{0.97,0.98,0.97}{3\%}\\
\hline Health & 30~\statbox{0.85,0.93,0.85}{15\%}\\
\hline Humanities & 41~\statbox{0.80,0.90,0.80}{21\%}\\
\hline Nutrition & 8~\statbox{0.96,0.98,0.96}{4\%}\\
\hline Simulation & 8~\statbox{0.96,0.98,0.96}{4\%}\\
\hline Social / Socioeconomic & 22~\statbox{0.89,0.95,0.89}{11\%}\\
\hline Other & 93~\statbox{0.53,0.77,0.53}{47\%}\\
\hline 
\hline \textbf{6.1.2. Target Variable} & \textbf{199}~\statbox{0.00,0.51,0.00}{100\%}\\ 
\hline Binary (categorical) & 39~\statbox{0.80,0.90,0.80}{20\%}\\
\hline Multi-class (categorical) & 128~\statbox{0.36,0.68,0.36}{64\%}\\
\hline Multi-label (categorical) & 9~\statbox{0.96,0.98,0.96}{5\%}\\
\hline Continuous (regression problems) & 24~\statbox{0.88,0.94,0.88}{12\%}\\
\hline Other & 38~\statbox{0.82,0.91,0.82}{19\%}\\
\hline 
\hline \textbf{6.2.1. ML Methods} & \textbf{198}~\statbox{0.01,0.51,0.01}{99\%}\\ 
\hline Convolutional Neural Network (CNN) & 25~\statbox{0.87,0.94,0.87}{13\%}\\
\hline Deep Convolutional Network (DCN) & 8~\statbox{0.96,0.98,0.96}{4\%}\\
\hline Deep Feed Forward (DFF) & 10~\statbox{0.95,0.98,0.95}{5\%}\\
\hline Deep Neural Network (DNN) & 19~\statbox{0.91,0.95,0.91}{10\%}\\
\hline Deep Q-Network (DQN)& 9~\statbox{0.96,0.98,0.96}{5\%}\\
\hline Generative Adversarial Network (GAN) & 10~\statbox{0.95,0.98,0.95}{5\%}\\
\hline Long Short-Term Memory (LSTM) & 13~\statbox{0.93,0.97,0.93}{7\%}\\
\hline Recurrent Neural Network (RNN) & 18~\statbox{0.91,0.96,0.91}{9\%}\\
\hline Variational Auto-Encoder (VAE) & 14~\statbox{0.93,0.96,0.93}{7\%}\\
\hline Other (DL methods) & 21~\statbox{0.89,0.95,0.89}{11\%}\\
\hline Linear (DR) & 57~\statbox{0.71,0.86,0.71}{29\%}\\
\hline Non-linear (DR) & 51~\statbox{0.75,0.87,0.75}{26\%}\\
\hline Bagging (ensemble learning) & 27~\statbox{0.87,0.93,0.87}{14\%}\\
\hline Boosting (ensemble learning) & 11~\statbox{0.95,0.97,0.95}{6\%}\\
\hline Stacking (ensemble learning) & 6~\statbox{0.97,0.98,0.97}{3\%}\\
\hline Other (generic)& 97~\statbox{0.51,0.76,0.51}{49\%}\\
\hline 
\hline \textbf{6.2.2. ML Types} & \textbf{197}~\statbox{0.02,0.51,0.02}{99\%}\\ 
\hline Classification (supervised)& 111~\statbox{0.44,0.73,0.44}{56\%}\\
\hline Regression (supervised) & 20~\statbox{0.90,0.95,0.90}{10\%}\\
\hline Other (supervised)& 7~\statbox{0.96,0.98,0.96}{4\%}\\
\hline Association (unsupervised) & 5~\statbox{0.98,0.99,0.98}{3\%}\\
\hline Clustering (unsupervised)& 41~\statbox{0.80,0.90,0.80}{21\%}\\
\hline Dimensionality Reduction (unsupervised)& 66~\statbox{0.67,0.84,0.67}{33\%}\\
\hline Classification (semi-supervised)& 13~\statbox{0.93,0.97,0.93}{7\%}\\
\hline Clustering (semi-supervised)& 6~\statbox{0.97,0.98,0.97}{3\%}\\
\hline Classification (reinforcement)& 1~\statbox{1.00,1.00,1.00}{1\%}\\
\hline Control (reinforcement)& 3~\statbox{0.98,0.99,0.98}{2\%}\\
\hline	
\end{tabular} 
\bigbreak
Legend: \thinspace
\statbox{1.00,1.00,1.00}{0\%} {\strut 0 papers} \thinspace
\statbox{0.49,0.75,0.49}{50\%} {\strut 100 papers} \thinspace
\statbox{0.00,0.51,0.00}{100\%} {\strut 200 papers} 
\end{minipage}
\hspace{1mm}
\begin{minipage}[t]{0.3\textwidth}
\begin{tabular}[t]{|l|r|}
\hline \textbf{6.3. ML Processing Phase} & \textbf{198}~\statbox{0.01,0.51,0.01}{99\%}\\ 
\hline Pre-processing / Input & 36~\statbox{0.82,0.91,0.82}{18\%}\\
\hline In-processing / Model & 45~\statbox{0.78,0.89,0.78}{23\%}\\
\hline Post-processing / Output & 162~\statbox{0.19,0.60,0.19}{81\%}\\
\hline
\hline \textbf{6.4. Treatment Method} & \textbf{196}~\statbox{0.02,0.51,0.02}{98\%}\\ 
\hline Model-agnostic / Black Box & 144~\statbox{0.28,0.64,0.28}{72\%}\\
\hline Model-specific / White Box & 70~\statbox{0.65,0.83,0.65}{35\%}\\
\hline 
\hline \textbf{6.5.1. Dimensionality} & \textbf{199}~\statbox{0.00,0.50,0.00}{100\%}\\ 
\hline 2D & 196~\statbox{0.02,0.51,0.02}{98\%}\\
\hline 3D & 5~\statbox{0.98,0.99,0.98}{3\%}\\
\hline
\hline \textbf{6.5.2. Visual Aspects} & \textbf{199}~\statbox{0.00,0.51,0.00}{100\%}\\ 
\hline Computed & 195~\statbox{0.02,0.51,0.02}{98\%}\\
\hline Mapped & 109~\statbox{0.45,0.73,0.45}{55\%}\\
\hline 
\hline \textbf{6.5.3. Visual Granularity} & \textbf{200}~\statbox{0.00,0.50,0.00}{100\%}\\ 
\hline Aggregated Information & 183~\statbox{0.09,0.55,0.09}{92\%}\\
\hline Instance-based / Individual & 146~\statbox{0.27,0.64,0.27}{73\%}\\
\hline 
\hline \textbf{6.5.4. Visual Representation} & \textbf{199}~\statbox{0.00,0.51,0.00}{100\%}\\ 
\hline Bar Charts & 82~\statbox{0.59,0.80,0.59}{41\%}\\
\hline Box Plots & 11~\statbox{0.95,0.97,0.95}{6\%}\\
\hline Matrix & 50~\statbox{0.75,0.87,0.75}{25\%}\\
\hline Glyphs / Icons / Thumbnails & 63~\statbox{0.69,0.84,0.69}{32\%}\\
\hline Grid-based Approaches & 19~\statbox{0.91,0.95,0.91}{10\%}\\
\hline Heatmaps & 46~\statbox{0.77,0.89,0.77}{23\%}\\
\hline Histograms & 56~\statbox{0.72,0.86,0.72}{28\%}\\
\hline Icicle Plots & 6~\statbox{0.97,0.98,0.97}{3\%}\\
\hline Line Charts & 56~\statbox{0.72,0.86,0.72}{28\%}\\
\hline Node-link Diagrams & 47~\statbox{0.76,0.88,0.76}{24\%}\\
\hline Parallel Coordinates Plots (PCPs) & 32~\statbox{0.84,0.92,0.84}{16\%}\\
\hline Pixel-based Approaches & 8~\statbox{0.96,0.98,0.96}{4\%}\\
\hline Radial Layouts & 22~\statbox{0.89,0.95,0.89}{11\%}\\
\hline Scatterplot Matrices (SPLOMs) & 18~\statbox{0.91,0.96,0.91}{9\%}\\
\hline Scatterplot / Projections & 115~\statbox{0.42,0.71,0.42}{58\%}\\
\hline Similarity Layouts & 27~\statbox{0.87,0.93,0.87}{14\%}\\
\hline Tables / Lists & 86~\statbox{0.57,0.78,0.57}{43\%}\\
\hline Treemaps & 5~\statbox{0.98,0.99,0.98}{3\%}\\
\hline Other & 59~\statbox{0.71,0.85,0.71}{30\%}\\
\hline 
\hline \textbf{6.5.5. Interaction Technique} & \textbf{185}~\statbox{0.07,0.54,0.07}{93\%}\\ 
\hline Select & 163~\statbox{0.92,0.96,0.92}{8\%}\\
\hline Explore / Browse & 169~\statbox{0.18,0.59,0.18}{82\%}\\
\hline Reconfigure & 74~\statbox{0.63,0.82,0.63}{37\%}\\
\hline Encode & 112~\statbox{0.44,0.72,0.44}{56\%}\\
\hline Filter / Query & 113~\statbox{0.44,0.72,0.44}{57\%}\\
\hline Abstract / Elaborate & 177~\statbox{0.42,0.71,0.42}{59\%}\\
\hline Connect & 128~\statbox{0.36,0.68,0.36}{64\%}\\
\hline Guide / Sheperd & 48~\statbox{0.76,0.88,0.76}{24\%}\\
\hline Verbalize & 9~\statbox{0.96,0.98,0.96}{5\%}\\
\hline 
\end{tabular} 
\end{minipage}
\hspace{1mm}
\begin{minipage}[t]{0.26\textwidth}
\begin{tabular}[t]{|l|r|}
\hline \textbf{6.5.6. Visual Variable} & \textbf{196}~\statbox{0.02,0.51,0.02}{98\%}\\ 
\hline Color & 195~\statbox{0.02,0.51,0.02}{98\%}\\
\hline Opacity & 83~\statbox{0.58,0.79,0.58}{42\%}\\
\hline Position / Orientation & 58~\statbox{0.71,0.85,0.71}{29\%}\\
\hline Shape & 37~\statbox{0.82,0.91,0.82}{19\%}\\
\hline Size & 68~\statbox{0.66,0.83,0.66}{34\%}\\
\hline Texture & 17~\statbox{0.91,0.96,0.91}{9\%}\\
\hline 
\hline \textbf{6.6. Evaluation} & \textbf{200}~\statbox{0.00,0.50,0.00}{100\%}\\ 
\hline Standard & 38~\statbox{0.81,0.91,0.81}{19\%}\\
\hline Comparative & 12~\statbox{0.94,0.97,0.94}{6\%}\\
\hline Before / During Development & 36~\statbox{0.82,0.91,0.82}{18\%}\\
\hline After Development & 47~\statbox{0.76,0.88,0.76}{24\%}\\
\hline Not Evaluated & 102~\statbox{0.49,0.75,0.49}{51\%}\\
\hline 	
\hline \textbf{6.7. Trust Levels (TL) 1--5} & \textbf{200}~\statbox{0.00,0.50,0.00}{100\%}\\ 
\hline Source Reliability & 11~\statbox{0.95,0.97,0.95}{6\%}\\
\hline Transparent Collection Process & 6~\statbox{0.97,0.98,0.97}{3\%}\\
\noalign{
\global\dimen1\arrayrulewidth
\global\arrayrulewidth1pt
}
\hline
\noalign{
\global\arrayrulewidth\dimen1 
}Uncertainty Awareness & 29~\statbox{0.85,0.93,0.85}{15\%}\\
\hline Equality / Data Bias & 16~\statbox{0.92,0.96,0.92}{8\%}\\
\hline Comparison (of Structures) & 86~\statbox{0.57,0.78,0.57}{43\%}\\
\hline Guidance / Recommendations & 46~\statbox{0.77,0.89,0.77}{23\%}\\
\hline Outlier Detection & 68~\statbox{0.66,0.83,0.66}{34\%}\\
\noalign{
\global\dimen1\arrayrulewidth
\global\arrayrulewidth1pt
}
\hline
\noalign{
\global\arrayrulewidth\dimen1 
}Familiarity & 3~\statbox{0.98,0.99,0.98}{2\%}\\
\hline Understanding / Explanation & 95~\statbox{0.53,0.76,0.53}{48\%}\\
\hline Debugging / Diagnosis & 54~\statbox{0.73,0.87,0.73}{27\%}\\
\hline Refinement / Steering & 69~\statbox{0.65,0.83,0.65}{35\%}\\
\hline Comparison & 61~\statbox{0.69,0.85,0.69}{31\%}\\
\hline Knowledgeability & 10~\statbox{0.95,0.98,0.95}{5\%}\\
\hline Fairness & 6~\statbox{0.97,0.98,0.97}{3\%}\\
\noalign{
\global\dimen1\arrayrulewidth
\global\arrayrulewidth1pt
}
\hline
\noalign{
\global\arrayrulewidth\dimen1 
}Experience & 7~\statbox{0.96,0.98,0.96}{4\%}\\
\hline In Situ Comparison & 54~\statbox{0.73,0.87,0.73}{27\%}\\
\hline Performance & 108~\statbox{0.46,0.73,0.46}{54\%}\\
\hline What-if Hypotheses & 40~\statbox{0.80,0.90,0.80}{20\%}\\
\hline Model Bias & 19~\statbox{0.91,0.95,0.91}{10\%}\\
\hline Model Variance & 16~\statbox{0.92,0.96,0.92}{8\%}\\
\noalign{
\global\dimen1\arrayrulewidth
\global\arrayrulewidth1pt
}
\hline
\noalign{
\global\arrayrulewidth\dimen1 
}Agreement of Colleagues & 9~\statbox{0.96,0.98,0.96}{5\%}\\
\hline Visualization Evaluation & 87~\statbox{0.56,0.78,0.56}{44\%}\\
\hline Metrics Validation / Results & 130~\statbox{0.35,0.67,0.35}{65\%}\\
\hline User Bias & 7~\statbox{0.96,0.98,0.96}{4\%}\\
\hline 
\hline \textbf{6.8. Target Group} & \textbf{196}~\statbox{0.02,0.51,0.02}{98\%}\\ 
\hline Beginners & 41~\statbox{0.80,0.90,0.80}{21\%}\\
\hline Practitioners / Domain Experts & 162~\statbox{0.19,0.60,0.19}{81\%}\\
\hline Developers & 36~\statbox{0.82,0.91,0.82}{18\%}\\
\hline ML Experts & 73~\statbox{0.64,0.82,0.64}{37\%}\\
\hline 	
\end{tabular} 
\end{minipage} 
\label{tab:categorizationLegend}
\vspace{-1.5em}
\end{table*}

\subsection{Data}

Many visualization techniques have been tested with specific data sets coming from different domains. However, just a few of them specifically work for one type of data set, for instance, the systems proposed by Bremm et al.~\cite{Bremm2011Assisted} and Wang et al.~\cite{Wang2019DeepVID}. In this subsection, we present the most frequent data domains we spotted and the nature of the target variable that should be predicted by ML classifiers.

\subsubsection{Domain}

DeepVID, by Wang et al.~\cite{Wang2019DeepVID}, is an example that focuses only on images (i.e., the overall field of \textbf{computer vision}). It shows how ML models for image classification can be interpreted and debugged with the use of simpler models (e.g., a linear model) in DNNs. Since DNNs usually work well with image data, exploring and diagnosing the training process of a DNN is an initial step towards boosting trust in them. 
With regard to \textbf{humanities} data, Sherkat et al.~\cite{Sherkat2018Interactive} propose a system that empowers users incorporating their feedback to define several diverse algorithms for clustering. The supported interactions enable users to adjust (or even create) new key terms, which are then used to supervise the algorithms and cluster the different documents. Involving humanities experts in user studies to evaluate the effectiveness of VA systems can further increase trust. 
An example of operating with \textbf{health} data is INFUSE~\cite{Krause2014INFUSE}, which helps analysts to select features and retrieve extra information based on a selection from a collection of algorithms, cross-validations folds, and different ML models. The visual representations assist domain experts (i.e., doctors) in manipulating their medical records more precisely and improve the accuracy of the results. In this case, the visualization tool seems to be generalizable to other domains. In medicine, receiving recommendations during the data processing phase is necessary to ensure that no further biases are introduced in the input phase of an ML model. This guidance is better achievable by feature exploration and feature selection with the use of visualization.
Bremm et al.~\cite{Bremm2011Assisted} focus on \textbf{biological} data sets. In their paper, the authors utilize scatterplot- and grid-based visualizations to facilitate selection and later comparison of data descriptors for unlabeled biology-related data. Except for the comparison of data and structures, any uncertainties stemming from the data should be highlighted for the biologists to focus on them. As a consequence, this extraction of patterns through visualization can increase their trust in ML. A number of papers focus on various other data domains, such as the works of Gleicher~\cite{Gleicher2013Explainers}, Sips et al.~\cite{Sips2009Selecting}, and Tatu et al.~\cite{Tatu2012Subspace}. 
%
\nocite{Alvarez2018On,Laugel2018Defining,Liu2014Distortion,Rieck2014Enhancing,Schreck2010Techniques,Turkay2016Enhancing}

\subsubsection{Target Variable}\label{sssection:category:target:variable}

In ML, the target (or response) variable is the characteristic known during the learning phase that has to be predicted for new data by the learned model. In classification problems, it can take a \emph{binary} value for two-class problems, a single label for \emph{multi-class} problems, or even a set of labels for \emph{multi-label} problems. In regression problems, it is generally a \emph{continuous} variable. 
On the one hand, Krause et al.~\cite{Krause2017AWorkflow} propose a workflow to help practitioners to examine, diagnose, and explain the decisions made by a binary classifier. In their approach, instance-level explanations are obtained based on local feature significance measures that explain single instances. With these findings as a basis, they develop visualizations that lead the users to important areas of investigation. Extensions of this approach could evaluate the reliability of the incoming data by comparing different areas and timeframes of the data.
On the other hand, a multi-class data set has been used in ActiVis~\cite{Kahng2018ActiVis}. This visualization tool integrates coordinated multiple views, such as a graph that provides an overview of the model architecture, and a neuron activation view for exploring DNN models with user-defined subsets of instances. To wrap up the decisions made based on the provided views, agreement of colleagues could further enhance the trust in these complex DNNs. 
Many approaches exist for regression problems, such as those described in the publications of Fernstad et al.~\cite{Fernstad2013Quality} and Hohman et al.~\cite{Hohman2019TeleGam}. For instance,  Piringer et al.~\cite{Piringer2010HyperMoVal} describe a validation framework for regression models that enable users to compare models and analyze regions with poor predictive performances. 
The optimization of the so-called \emph{trade-off of bias and variance} is also crucial for regression problems.
%
\nocite{Hollt2019Focus,May2011Guiding,Wang2018The,Zhao2019FeatureExplorer}

Papers categorized as \emph{others} on the target variable group concern ML settings in which no target variable is available. These are mostly related to DR and clustering problems. 
The method designed by Zhou et al.~\cite{Zhou2016Dimension}, for example, combines both aspects to enable users to design new dimensions from data projections of subspaces, with the goal of maintaining important cluster information. The newly adapted dimensions are included in the analysis together with the original ones, to help users in forming target-oriented subspaces that explain---as much as possible---cluster structures.

\subsection{Machine Learning}

This subsection covers various ML methods that were divided into three main classes: \emph{DL}, \emph{DR}, and \emph{ensemble learning}. We also discuss different ML types that we considered in our categorization:  \emph{supervised learning}, \emph{unsupervised learning}, \emph{semi-supervised learning}, and \emph {reinforcement learning}.

\subsubsection{Machine Learning Methods}

In the area of \emph{DL}, we observed two categories that are in the focus of most DL-related papers: \emph{CNNs} and \emph{RNNs}. 
CNNComparator~\cite{Zeng2017CNNComparator} addresses the challenges of comparing CNNs and enables users to freeze the tool for different epochs of a trained CNN model (by using so-called snapshots). An epoch is completed when a data set is processed one time forward and backward through an NN. Thus, CNNComparator provides insights into the architectural design, and as a consequence, it enables better training of CNN models. 
For RNNs, RNNbow~\cite{Cashman2018RNNbow} visualizes the gradient flow while backpropagation occurs in the training of RNNs. By visualizing the gradient, this tool offers insights into how exactly the network is learning. Both  papers explicitly enhance trust in different DL models by either comparing CNNs or explaining RNNs to the users via visualization.
In the \emph{DR} subclass, linear techniques surpass non-linear ones in volume (the former was found in 57 papers $vs.$ 51 for the latter). One example here is the iPCA tool~\cite{Jeong2009iPCA}. It augments the principal component analysis (PCA) algorithm with an interactive visualization system that supports the investigation of relationships between the data and the computed eigenspace. 
Overall, the tool employs views for exploring the data, the projections, the PCA's eigenvectors, and the correlations between them. Eventually, prominent uncertainties become aware to users by examining all these relations.
In another example, AxiSketcher~\cite{Kwon2017AxiSketcher} enables users to impose their domain knowledge on a visualization by allowing interaction using a direct-manipulation technique over a t-SNE projection (non-linear DR technique). Users can sketch lines over specific data points, and the system composes new axes that represent a non-linear and weighted mixture of multidimensional attributes. Thus, the comparison of clusters enables users to identify problematic cases (in terms of trust) in a projection. 
For \emph{ensemble learning}, \emph{bagging} is the most common category. iForest~\cite{Zhao2019iForest} is a visualization tool that accommodates users with an aggregated view showing and summarizing the decision paths in random forests, which finally reveals the working mechanism of the ML model. Visualizing and understanding the decision paths of random forest algorithms, as well as how their performance was reached, serves as a foundation for assessing the trust in bagging ensemble learning. 
Other, more general examples can be found in the works by Schneider et al.~\cite{Schneider2018Integrating} and Sehgal et al.~\cite{Sehgal2018Visual}.

\subsubsection{Machine Learning Types}

According to our analysis, \emph{supervised learning} and \emph{classification} problems are extensively addressed by the visualization community. For instance, a visualization system that works with choosing the best classifiers is EnsembleMatrix~\cite{Talbot2009EnsembleMatrix}. It allows users to directly interact with the visualizations in order to explore and build combinations of models. Comparison of ML models and validation metrics are key factors in increasing trust in them. HyperMoVal~\cite{Piringer2010HyperMoVal}, already discussed in Section \ref{sssection:category:target:variable}, focuses on \emph{regression} problems and provides several functionalities: comparing the ground truth against predicted results, analyzing areas with a poor fit, evaluating the physical plausibility of models, and comparing various classifiers. When users address regression problems, the comparison of alternative ML models and steering each of them can also improve the trustworthiness of the models. In \emph{unsupervised learning}, a \emph{clustering} example is the visualization technique developed by Turkay et al.~\cite{Turkay2011Interactive}, which visualizes the structural quality of several temporal clusters at a certain point in time or over time. DimStiller~\cite{Ingram2010DimStiller} is a system (belonging to the \emph{DR} subclass) that assists the user in converting the input dimensions in a number of analytical steps into data tables that can be transformed into each other with the help of so-called operators. Users can manipulate those operators for parameter tuning and for guidance to discover patterns in the local neighborhood of the data space. Both DR and clustering visualization tools often utilize comparison of structures and emphasize patterns observable in projections. Some rare cases are related to \emph{semi-supervised learning}, such as MacInnes et al.~\cite{MacInnes2010Visual} and Bernard et al.~\cite{Bernard2018Towards}. \emph{Reinforcement learning} is covered by the work of Saldanha et al.~\cite{Saldanha2019ReLVis}, for instance. 

\subsection{Machine Learning Processing Phase}

VASSL~\cite{Khayat2019VASSL} is a system that works with the \emph{pre-processing/input phase} and enhances the performance and scalability of the manual labeling process by providing multiple coordinated views and utilizing DR, sentiment analysis, and topic modeling. The system allows users to select and further investigate batches of accounts, which supports the discovery of spambot cases that may not be detected when checked independently. For the \emph{in-processing/model phase}, Liu et al.~\cite{Liu2017Towards} designed a tool that helps to better understand, diagnose, and steer deep CNNs. They represent a deep CNN as a directed acyclic graph, and based on this representation, a hybrid visualization has been developed to disclose multiple aspects of each neuron and the intercommunications between them. The largest category with regard to the number of available visualizations is \emph{post-processing/output} for visualizing the final results, such as MultiClusterTree~\cite{VanLong2009MultiClusterTree}. In their tool, the authors propose a 2D radial layout that supports an inherent understanding of the distribution arrangement of a multidimensional multivariate data set. Unique clusters can be explored interactively by using parallel coordinates when being selected in a cluster tree representation. The overall cluster distribution can be explored, and better understanding of the relations between clusters and the initial attributes is supported as well. As expected, the input phase is highly related to TL2, the in-processing phase to understanding and steering categories of TL3, and the final phase to metrics validation (TL5). Finally, Gil et al.~\cite{Gil2019Towards} and Sacha et al. with VIS4ML~\cite{Sacha2019VIS4ML} are two workflow papers that provide an overview of all these phases.

\subsection{Treatment Method}

\emph{Model-agnostic} techniques are twice as common as \emph{model-specific} techniques. With the former, we mean---in most of the cases---visualization methods that treat ML models as \emph{black boxes}. The latter is usually connected to techniques specifically developed to open these black boxes, and thus make the ML models to be regarded as \emph{white boxes}. An example of a model-agnostic visualization tool is ATMSeer~\cite{Wang2019ATMSeer} with which users are able to steer the search space of AutoML and explain the results. A multi-granular visualization empowers users to observe the AutoML process, examine the explored ML models, and refine the search space in realtime. 
In the white box case, the visualization tool EasySVM~\cite{Ma2017EasySVM} facilitates users in tuning parameters, managing the training data, and extracting rules as a component of the support vector machine (SVM) training process. The goal of model-specific techniques is to explain the inner workings of a particular ML model. However, some tools combine both specific models and model-agnostic algorithms, such as Chae et al.~\cite{Chae2017Visualization}, Roesch and G{\"u}nther~\cite{Roesch2019Visualization}, Pezzotti et al.~\cite{Pezzotti2018DeepEyes}, and others~\cite{Caballero2019V,Kinkeldey2019Towards,Martins2014Visual}.

\subsection{Visualization}

Various approaches, types, and properties of visualization are used in our 200 surveyed papers, often in combinations. The knowledge of the most common techniques and approaches can guide early-stage researches to choose the most important of them or senior researchers to discover potential gaps in the literature. The selection of the best visualization approaches, types, and properties for a given situation can effectively reduce potential visualization bias. Successfully addressing questions such as ``where, when, and why should I use a 2D bar chart to present aggregated information instead of another visual representation?'', for instance, can boost trust in ML models. Carefully thinking about which data should be visualized is similarly important. This section of our report describes all these aspects and introduces the corresponding category groups.

\subsubsection{Dimensionality}

With regard to dimensionality of the visual display, almost all visualizations (196) are \emph{2D}, such as~\cite{Amorim2012iLAMP,Johansson2009Interactive,Ming2019ProtoSteer,Stahnke2016Probing}. An exception is the interactive visualization technique by Coimbra et al.~\cite{Coimbra2016Explaining} that adapts and improves biplots to show the data attributes in the projected \emph{three-dimensional (3D)} space. They use interactive bar chart legends to present variables that are visible from a given angle and also support users to decide on the optimal position to examine a desired set of attributes. 

\subsubsection{Visual Aspects}
The information to be visualized can either be directly mapped from the data values themselves or be computed (algorithmically derived). ModelTracker~\cite{Amershi2015ModelTracker} extracts information contained in conventional summary statistics and charts while letting users examine errors and diagnose ML models. Hence, it contains \emph{computed} instead of \emph{mapped} instances. Arendt et al.~\cite{Arendt2019Towards} visualize the classifier's feedback after each iteration with their IML interface. To address scalability issues of the visualization, this interface communicates with the user by a small set of system-proposed instances for each class.

\subsubsection{Visual Granularity}

G{\"o}rtler et al.~\cite{Gortler2020Uncertainty} represent \emph{aggregated information} in their visualizations. They use a technique that performs DR on data that is subject to uncertainty by using a generalization of standard PCA. Their technique helps to discover high-dimensional characteristics of probability distributions and also supports sensitivity analysis on the uncertainty in the data. Zeiler and Fergus~\cite{Zeiler2014Visualizing} introduce a visualization technique that contributes to insight generation for the general operation of the classifier in an \emph{instance-based} manner, i.e., for individual data cases. 
Nevertheless, most visualization systems and techniques involve both the exploration of aggregated information and individual cases, e.g., presented by Choo et al.~\cite{Choo2010iVisClassifier}  and the visualization tool BaobabView~\cite{VanDenElzen2011BaobabView}.

\subsubsection{Visual Representation}

Liu et al.~\cite{Liu2018Visual} combine multiple coordinated views to provide a thorough overview of a tree boosting model and enable the effective debugging of a failed training process. One of their views utilizes \emph{bar charts} in order to rank the most valuable features that affect the model's performance. Ji et al.~\cite{Ji2019Visual} propose visual exploration of a neural document embedding with the goal to gain insights into the underlying embedding space and encourage this utilization in standard infrared (IR) spectroscopy applications. In their paper, they use a \emph{scatterplot} visualization, i.e., a projection. LSAView~\cite{Crossno2009LSAView} is a system for interactive, latent semantic analysis (LSA) models. Multiple views, linked matrix-graph views, and data views in the form of \emph{lists} are used to choose parameters and see the effects of them. Other papers apply different visual representations, some rare cases are \emph{waterfall charts}, \emph{violin charts}, \emph{Voronoi diagrams}, and \emph{bipartite graphs}~\cite{Aupetit2007Visualizing,Hohman2019TeleGam,Khayat2019VASSL,Lespinats2011CheckViz,Lin2018RCLens,Wang2019DQNViz,Wang2018GANViz,Zhao2016Manifold}.

\subsubsection{Interaction Technique}

Gehrmann et al.~\cite{Gehrmann2019Visual} argue that both the visual interface and model architecture of DL systems need to consider the interaction design. They propose a collaborative semantic inference for the constructive cooperation between humans and algorithms. Semantic interactions permit a user both to understand and regulate parts of a model's reasoning process. All these interactions enable the \emph{selection} of particular sentences and then further \emph{exploration} of the content with suggestions stemming from the system side. \emph{Abstract/elaborate} is another interaction technique found in, e.g., Borland et al.~\cite{Borland2019Selection} and can be interpreted as different granularities that the visualization allows users to explore the data. 
Sevastjanova et al.~\cite{Sevastjanova2018Going} argue that a combination of visualization and \emph{verbalization} methods is advantageous for generating wide and versatile insights into the structure and decision-making processes of ML models. For more details about the remaining interaction techniques (e.g.,~\cite{Cavallo2018Track,Chegini2019Interactive,Choo2010iVisClassifier,Padua2014Interactive,Park2019ComDia,Zhao2018SkyLens}), we refer to the survey of Lu et al.~\cite{Lu2017The}.

\subsubsection{Visual Variable}

Ahmed et al.~\cite{Ahmed2011Steerable} use a qualitative \emph{color} scheme in order to encode cluster groupings in all views of their visualization tool for steering mixed-dimensional KD-KMeans clustering. Color is used in almost every paper we examined~\cite{Chegini2018Interactive,ElAssady2018ThreadRecostructor,Kienreich2012Visual,Molchanov2014Interactive,Peltonen2017Negative,Xia2016DimScanner}. DeepCompare~\cite{Murugesan2019DeepCompare} uses \emph{opacity} and \emph{size}, which are the two second-most occurring \emph{visual variables}. Their tool visualizes the results of DL models, provides insights into the model behavior and the assessment of trade-offs between two such models. In more detail, the activation value of an NN is encoded as \emph{size}, while \emph{opacity} is used to remove the highlighting when specific cases are selected.

\subsection{Evaluation}\label{sec:eval}

In this subsection, we explore how visualizations are evaluated in our community and how many of them had been actually evaluated. Surprisingly, around half of the visualizations were never evaluated. The evaluation of visualizations is a fundamental component to validate the usability of visualization tools and systems.
%
\nocite{Cutura2018VisCoDeR,Das2019BEAMES,Krause2016Using,Paiva2012Semi,Rauber2018Projections}

\subsubsection{User Study}

RuleMatrix~\cite{Ming2019RuleMatrix} is one of the approaches that follows a \emph{standard} procedure for performing an evaluation in the InfoVis community. That is, various participants had to solve a series of tasks by using the tool during which the accuracy and timing was monitored to gain insight into the usability of the proposed solution. The paper presents an interactive visualization technique to assist novice users of ML to understand, examine, and verify the performance of predictive models. 
FairSight~\cite{Ahn2019FairSight} is another tool designed to accomplish different concepts of fairness in ranking decisions. To achieve that, FairSight distinguishes the required actions (understanding, computing, and others) that can possibly lead to fairer decision making. It was \emph{compared} against the What-If Tool~\cite{Wexler2019The} and found to perform better and result in more benefits than the latter approach.
\subsubsection{User Feedback}

Cashman et al.~\cite{Cashman2019AUser} worked with exploratory model analysis, which is defined as the process of finding and picking relevant models that can be used to create predictions on a data source. \emph{During development}, they improved their tool and received user \emph{feedback}. Hazarika et al.~\cite{Hazarika2019NNVA} used networks as surrogate models for visual analysis, and \emph{after the development} of their system and techniques, a domain expert gave them feedback in order to further improve the VA system at the end of the development process. Ultimately, from the further analysis of the statistics, we conclude that in five cases both domain and ML experts used visualization tools and evaluated them. In 32 cases, e.g., \cite{Chuang2014Interactive,Kwon2019RetainVis,Liu2019NLIZE,Migut2011Interactive,Xia2018LDSScanner}, only domain experts were asked; and in 19 cases only ML experts  participated, such as in~\cite{Liu2018AnalyzingtheT,Nie2018Visualizing}.

\subsubsection{Not Evaluated}
As described earlier, approximately half of the papers did not include any type of evaluation. However, we discovered one visualization tool~\cite{Kahng2019GAN} that was evaluated later in a new publication~\cite{Kahng2019How}.

\subsection{Trust Levels}

The most novel components of the categorization presented in this section are the different levels of trust we identified in the 200 individual papers. In \autoref{sec:back}, we divided the enhancement of trust in ML models with the help of visualizations into five levels (raw and processed data, learning method, concrete model, and user expectation).  

\subsubsection{Raw Data (TL1)}

\emph{Source reliability} often comes together with \emph{transparent collection processes} as in AnchorViz~\cite{Chen2018AnchorViz}, an interactive visualization that facilitates erroneous regions detection through semantic data exploration. By pinning anchors on top of the visualization, users create a topology to lie upon data instances based on their relation to those nearby anchors. Examination of discrepancies between semantically related data points is another functionality of the tool. This data exploration helps to observe source reliability and if any strange effects occurred when the collection process happened. However, as can be seen from the data in Table~\ref{tab:categorizationLegend}, these two categories are covered rarely by visualization tools.

\subsubsection{Processed Data (TL2)}

\emph{Uncertainty awareness} and investigation is an established subject of research in the visualization community with techniques such as the one presented by Berger et al.~\cite{Berger2011Uncertainty}. The authors developed techniques that guide the user to potentially interesting parameter areas and visualize the intrinsic uncertainty of predictions in 2D scatterplots and parallel coordinates. FairVis~\cite{Cabrera2019FairVis} is a recent paper that addresses a new problem which seems to become a trend. \emph{Data bias and equality} is a major issue and should be---as much as possible---removed from our ML models. FairVis facilitates users to review the fairness of ML models in interesting, explored subgroups. iVisClustering~\cite{Lee2012iVisClustering} is one of the many visualization tools that allow the \emph{comparison of different structures} (clusters in this case) and \emph{guide/recommend} the users by proposing new clusters based on previous actions. With the help of such visualizations, users can interactively refine clustering results in various ways. Also, iVisClustering can fade away  noisy data and re-cluster the data accordingly to demonstrate a meaning representation. Zhao et al.~\cite{Zhao2019Oui} developed a tool that enables users to recognize, explain, and choose  outliers discovered by various algorithms. Roughly one third of the papers cover outlier detection related topics (68 out of 200), such as ~\cite{Liu2015Visual,Rauber2017Visualizing,Sacha2018SOMFlow}.

\subsubsection{Learning Method (TL3)}

The work of Olah et al.~\cite{Olah2018The} tries to \emph{familiarize} users with different DL algorithms. They test robust interfaces that appear when users appropriately combine them and the rich composition of this combinatorial solution space. In our STAR, interpretability and explainability is separated into four categories: \emph{understanding/explanation}, \emph{debugging/diagnosis}, \emph{refinement/steering}, and \emph{comparison}. These categories may even occur in pairs or triplets, depending on the visualization system and technique. For instance, Liu et al.~\cite{Liu2018Analyzing} support  \emph{understanding/explanation} of the reasons behind faulty predictions introduced by adversarial attack examples. The basic concept is to analyze groups of critical neurons and their connections of the adversarial attacks and match them with those of the normal cases. DeepTracker~\cite{Liu2018DeepTracker} facilitates the exploration of the intense dynamics of CNN training processes and helps to identify the unique patterns that are ``buried'' inside the enormous amount of information in a training log (\emph{debugging process}). Hamid et al.~\cite{Hamid2019Visual} describe a visual ensemble analysis based on hyper-parameter space and performance visualizations. These visualizations are mutually used with associations' explorations between topological arrangements and allow the production of enough knowledge in order to support users \emph{steering} the process. Rieck and Leitte~\cite{Rieck2015Comparing} suggest a \emph{comparative} analysis of DR methods according to what level of preservation of structural features in the high-dimensional space remains in the 2D embeddings. Local and global structural features are assessed in the original space, and specific DR methods are chosen based on those findings. Manifold~\cite{Zhang2019Manifold} is conceived of a generic framework that does not rely on the internals of particular ML models and only observes the input and the output. With the comparison of various models and learning methods, it allows users to become \emph{knowledgeable} about their usability. FairSight~\cite{Ahn2019FairSight} which was discussed earlier along with the What-If Tool~\cite{Wexler2019The} are both two recent examples of how \emph{fairness} is a trending unexplored subject in the community. The What-If Tool (which was not discussed yet) enables domain experts to assess the performance of models in hypothetical scenarios, analyze the significance of several data features, and visualize model functionality across many ML models and batches of input data. It also engages practitioners in grading systems that are able to show multiple ML fairness validation metrics.

\subsubsection{Concrete Model (TL4)}

Cashman et al.~\cite{Cashman2019Ablate} researched the rapid exploration of model architectures and parameters. To this end, they developed a VA tool that allows a model developer to discover a DL model immediately via exploration as well as rapid deployment and examination of NN architectures. By visually comparing models, beginners might come to similar conclusions (e.g., that early stages of convolutional layers perform well in feature extraction) as ML experts who take advantage of their \emph{experience}. \emph{In situ comparison}, i.e., a comparison of two or more states of the same model, is performed by Gamut~\cite{Hohman2019Gamut}, for example. 
The benefit of Gamut lies in the justification of why and how professional data scientists interpret models and what they look for when comparing their internal components. Our investigation showed that interpretability is not a monolithic concept: data scientists have different reasons to interpret models and tailor explanations for specific audiences, often balancing competing concerns of simplicity and completeness. Moreover, \emph{performance} is one of the most common techniques to choose from (see Table~\ref{tab:categorizationLegend}) when having different models. LoVis~\cite{Zhao2014LoVis} allows the user to progressively construct and validate models that promote local pattern discovery and summarization based on ``complementarity'', ``diversity'', and ``representativity'' of models. \emph{What-if hypotheses} are supported by Clustrophile~2~\cite{Cavallo2019Clustrophile2}, which guides users in a clustering-based exploratory analysis. It also adapts incoming user feedback to improve user recommendations, helps the interpretation of clusters, and supports the rationalization of differences between clusterings. Last but not least, papers that deal with issues related to \emph{model bias} and \emph{model variance} usually occur together. M{\"u}hlbacher and Piringer~\cite{Muhlbacher2013APartition} present a framework for building regression models addressing these limitations. Analyzing prediction bias with model residuals is one of the techniques used to limit the local prediction bias of a model, i.e., avoiding the inclination towards underestimation or overestimation. They also visualize the point-wise variance of the predictions by using a pixel-based view.

\subsubsection{Evaluation/User Expectation (TL5)}

The \emph{Agreement of colleagues} is related to provenance and the possibility to enable users to collaborate with each other. Wongsuphasawat et al.~\cite{Wongsuphasawat2018Visualizing} present a design study of the TensorFlow Graph Visualizer, which is a module of the shareable TensorFlow platform. This tool improves users' understanding of complicated ML architectures by visualizing data-flow graphs. These flows can be investigated, and at each point in time, provenance can be considered as a way to return back to a previous situation. 
\emph{Visualization evaluation}, as mentioned earlier in Sect.~\ref{sec:eval}, is activated when the visualization techniques and tools are evaluated or if any type of feedback is provided. Showing \emph{metrics validation/results} is the most common way of enhancing trust until now. Squares~\cite{Ren2017Squares} is a performance visualization for multi-class classification problems. Squares supports estimating standard \emph{performance metrics} while demonstrating instance-based distribution information essential for supporting domain experts in prioritizing efforts. Furthermore, Fujiwara et al.~\cite{Fujiwara2019Supporting} implemented a VA method that highlights the crucial dimensions of a cluster in a DR result. To obtain the important dimensions, they introduce an improved method of contrastive PCA. The method utilized is called ccPCA (contrasting clusters in PCA) and can compute each dimension's relevant contribution to one versus other clusters. An example that implicitly checks \emph{user bias} is the explAIner tool by Spinner et al.~\cite{Spinner2019explAIner}. explAIner is a VA system based on a framework that connects an iterative explainable ML pipeline with 8 global observing and refinement mechanisms, including ``quality monitoring'', ``provenance tracking'', or ``trust building''. Additionally, Jentner et al.~\cite{Jentner2018Minions} propose the metaphorical narrative methodology to translate mental models of the involved modeling and domain experts to machine commands and vice versa. The authors provide a human-machine interface and discuss crucial features, characteristics, and pitfalls of their approach. With regard to user bias, the research community has taken ``small steps''  with only a few papers tackling this issue. However, explicit reports about this challenge are still rare, unfortunately.

\subsection{Target Group}

In most cases, the visualization tools cover at least the \emph{target group} of \emph{domain experts/practitioners}~\cite{Engel2012Visual,Frohler2016GEMSe,Fujiwara2020An,Garg2008Model,Hoferlin2012Inter,Krause2016Interacting}. Then, other target groups such as \emph{ML experts}~\cite{Jiang2017Interactive,Kauer2018Mapping,Seifert2010Stress,Wang2017Linear} and \emph{developers} are in the focus of the authors~\cite{Kahng2016Visual,Madsen2019Visualizing,Rieck2015Persistent,Yan2018Homology} (commonly together).  \emph{Beginners/novice users}~\cite{Janik2019Interpreting,Ming2019Interpretable,Silva2015Attribute,Thiagarajan2018Exploring} are rarely considered. To give two examples, B{\"o}gl et al.~\cite{Bogl2014Visual} support with TiMoVA-Predict several types of predictions with a holistic VA approach that focuses on domain experts. Providing different prediction capabilities allows for assessing the predictions during the model selection process via an interactive visual environment. Biologists and doctors, for instance, are interested in being able to compare data structures and receive guidance on where to focus on. Ma et al.~\cite{Ma2019Explaining} employ a multi-faceted visualization schema intended to aid the analysis of ML experts for the domain of adversarial attacks.

\section{Survey Data Analysis} \label{sec:topic}
	\begin{table*}[htb!]
 \centering 
 \caption{For each of the ten topics, we present the top eight terms that we extracted from the results of the latent Dirichlet allocation (LDA) that has been applied to all papers. The suggested topic titles are shown in italics. Each topic is encoded by one specific color. In each topic, we cite the papers that mostly belong to them.}
 \includegraphics[width=\textwidth]{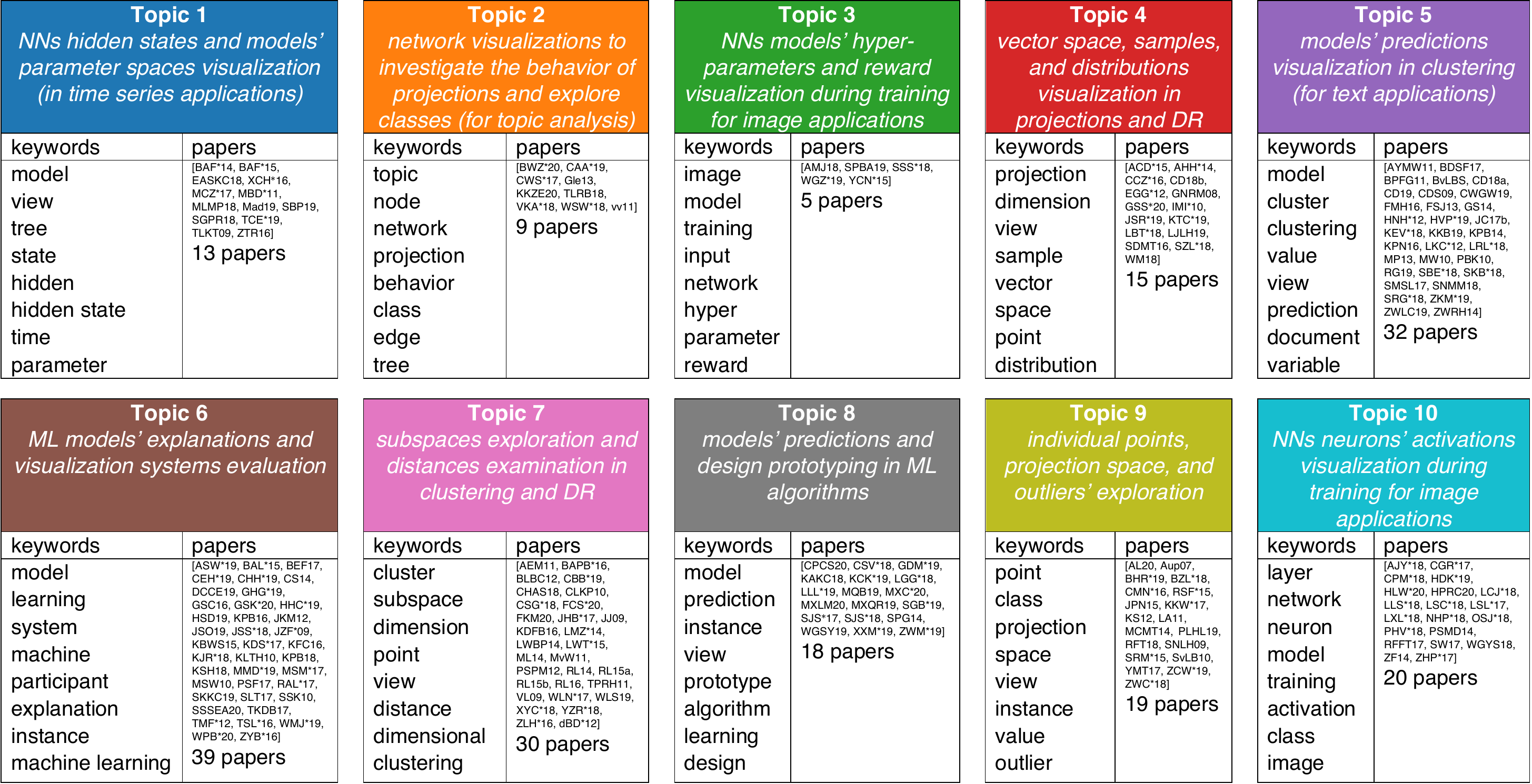}
 \label{tab:topics}
 %
 %
\end{table*}

In the previous parts of our report, we explained our overall methodology, provided high-level statistical information on the selected papers, and introduced our categorization together with example papers assigned to the individual categories. Now in this section, we discuss lower-level analytical results based on the collected papers and their metadata. 
In order to detect interesting connections and important emerging topics among the 200 papers, we applied topic modeling to all of them, following the visual text analysis approach by Kucher et al.~\cite{Kucher2018Analysis}. While the topic modeling results might be subject to the algorithm and parameter choice concerns to some extent, they provide  information complementary to the results of our manual investigations. Thus, the topic analysis contributes both to the validation and the new insights regarding the categorized publications. In addition, we investigate the relations between categories in general (again following the workflow proposed by Kucher et al.~\cite{Kucher2018The}) and explore the different data sets used in the individual papers. All those analyses help us to validate and further explore our categorization by creating new insights that can be used as research opportunities for this subject (cf. \autoref{sec:discuss}).

\subsection{Topic Analysis}

\begin{figure*}[htb!]
 \centering 
 \includegraphics[width=\textwidth]{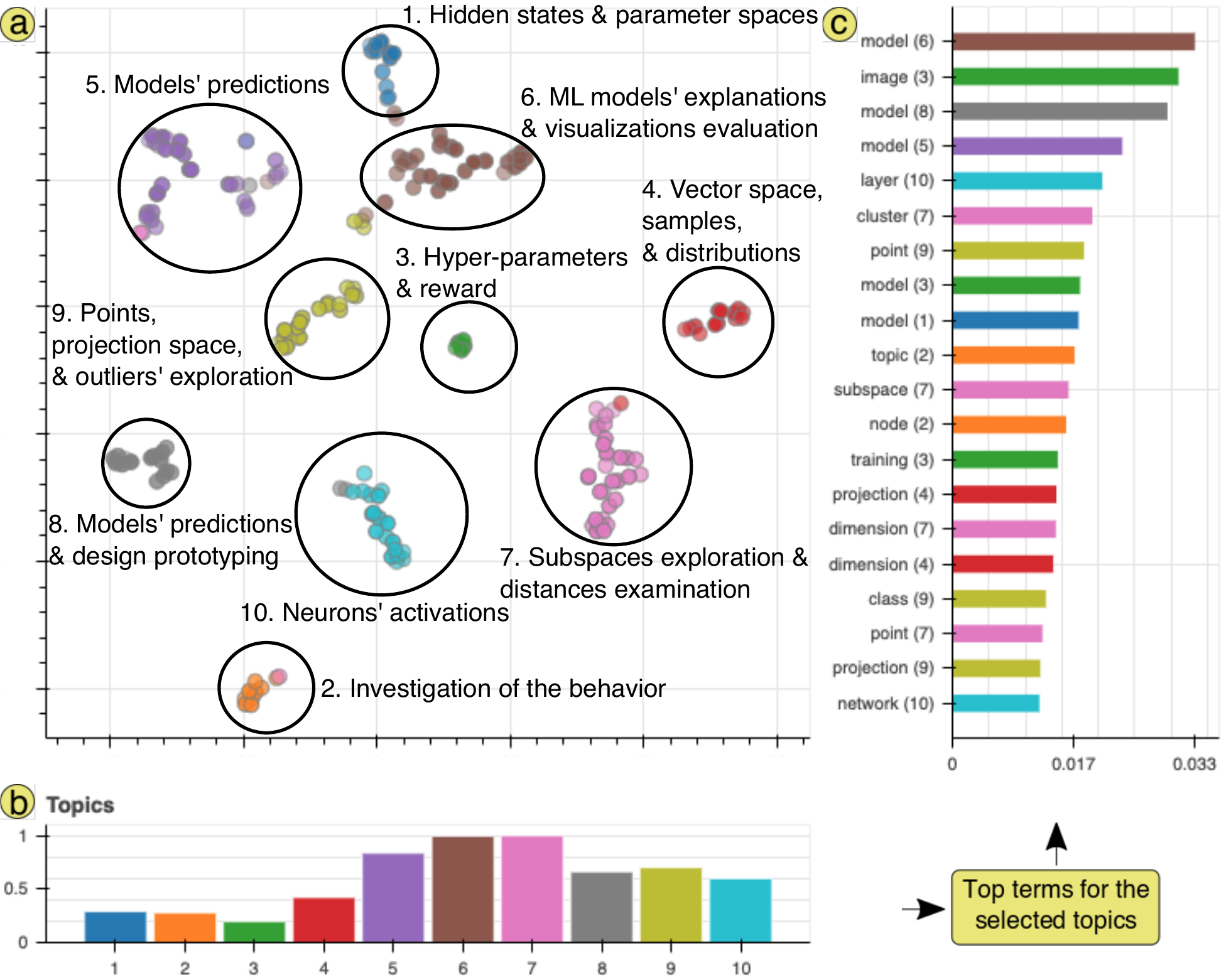}
 \caption{Visual exploration of new interesting topics derived from the 200 papers. (a) Papers' embedding generated with the t-SNE algorithm and based on the corresponding topics. The black outlines were manually drawn on top, and the tags act as short versions of the 10 full topic titles. (b) Bar chart of topics with each topic's significance (scaled from 0 to 1). (c) Horizontal bar chart of top terms with the highest relevance for all the topics. 
 Here, topics are encoded with color and number (in parentheses); a single term can be found in several topics.
 }
 \label{fig:topics_proj}
 \vspace{-.5em}
\end{figure*}

\noindent \textbf{Methodology.} \quad
First, we collected the PDF files of the selected papers and converted them to plain text. After that step, we prepared the text corpus by clearing the full texts from the authorship details and acknowledgments. Next, we processed them with the latent Dirichlet allocation (LDA) algorithm~\cite{Blei2003Latent,Griffiths2004Finding} (a common approach for topic modeling). In order to verify the LDA results---because it might produce diverse results at different executions---we ran the same process several times to get comparable results. The results do not indicate a major deviation from the main topic of each paper previously assigned by the manual categorization process. 
Finally, our LDA results led to ten topics (limited by us due to the lack of space and our attempt to choose a reasonable number of topics). 
The top eight terms for each topic are displayed in \autoref{tab:topics} together with the papers belonging to a topic (see \textbf{S6} for further details). From the terms that occurred, we removed any terms related to the structure of the analyzed texts and not to the actual content, for instance, ``figure'' and ``fig''. Our implementation is based on Python with NLTK~\cite{Bird2006NLTK} for the pre-processing of stop words and Gensim~\cite{Rehurek2010} for the topic modeling part. The names of the topics were manually assigned by us after several discussion cycles considering both the top terms and the contents of the papers in each topic. The results are then visualized with the assistance from the interactive visualization tool described by Kucher et al.~\cite{Kucher2018Analysis}, see \autoref{fig:topics_proj}. This visualization is based on a DR projection which may not be the most reliable approach. However, the ground truth labels taken from the LDA results match in almost all the cases with the clusters formed by the embedding.

\noindent \textbf{Topics.} \quad
In the following description list, we shortly summarize the ten topics (see Table~\ref{tab:topics}) we identified:
\begin{description}
  \item[\textcolor{ColT1}{\rule{\boxh}{\boxh}} Topic 1 -- hidden states \& parameter spaces.] \quad  
  According to our analysis, the common factor between the majority of the 13 papers in this topic class is their focus on time series data~\cite{Bogl2014Visual,Bogl2015Integrating} and RNNs~\cite{Sawatzky2019Visualizing} (in Strobelt et al.~\cite{Strobelt2018LSTMVis}: long short-term memory networks). It seems that the exploration of the hidden states of such networks preserves lost information that could enhance trust~\cite{Ming2017Understanding} with appropriate expert intervention. 
 Another subtopic here is the ML models' parameter spaces exploration~\cite{Tyagi2019ICE}, which enables users to find the best parameters based on a series of optimizations for certain goals. In this context, M{\"u}hlbacher et al.~\cite{Muhlbacher2018TreePOD} present an approach that visualizes the effects of these parameters. 
 As stated by these previous works, support for the visual parameter search is still an open research challenge. 
 \linebreak
  \item[\textcolor{ColT2}{\rule{\boxh}{\boxh}} Topic 2 -- investigation of the behavior.] \quad 
  This topic class contains 2 out of 9 papers on topic analysis applications~\cite{Chen2019LDA,Khayat2019VASSL}. 
  A common theme here is network visualization used for explaining Bayesian networks~\cite{Cypko2017Visual,Vogogias2018BayesPiles} and decision trees~\cite{VanDenElzen2011BaobabView}. 
  Other subtopics (which lead to research opportunities) are the exploration of behavior with regard to the decomposition of projections, showing the internal parts of ML models (and how classes are formed inside them), and the role of the user; they are all covered by our categorization presented in~\autoref{sec:categ}. 
  \item[\textcolor{ColT3}{\rule{\boxh}{\boxh}} Topic 3 -- hyper-parameters \& reward.] \quad 
  All five papers in this class (except one~\cite{Saldanha2019ReLVis} related to reinforcement learning) make use of image data. 
  They form a tight, green cluster in \autoref{fig:topics_proj} and focus mainly on DL techniques~\cite{Yosinski2015Understanding}. 
  It is an open challenge to decide which hyper-parameter value~\cite{Schall2018Visualization,Wang2019DeepVID} is better for a particular NN and specific problem, thus implying novel visualization techniques. 
  For reinforcement learning, instead, more research is needed for studying what behaviors are associated with the two types of reward (high and low) and monitoring how they develop during training. 
  \item[\textcolor{ColT4}{\rule{\boxh}{\boxh}} Topic 4 -- vector space, samples, \& distributions.] \quad 
  \sloppy In this topic class, three papers~\cite{Liu2018VisualExploration,Liu2019Latent,Sun2018FraudVis} work with vector space embeddings that illustrate similarities of the data. Most of the 15 papers highlight the importance of visualizing instances and samples~\cite{Kahng2019GAN,Stahnke2016Probing} that form clusters in projections and explore the distributions~\cite{Alsallakh2014Visual,Chen2016DropoutSeer,Cavallo2018AVisual} of points in DR techniques. Finding ways to improve these visualizations is still an open challenge in the InfoVis community. 
  \item[\textcolor{ColT5}{\rule{\boxh}{\boxh}} Topic 5 -- models' predictions.] \quad 
   Models' predictions and results visualization with the use of \emph{quality and validation metrics}~\cite{Fernstad2013Quality,Gotz2014Visualizing} (depending on the ML type) composes a big, more general topic class with 32 surveyed papers. A subgroup in this class especially refers to clustering challenges~\cite{Bernard2017Combining,Kwon2018Clustervision} and open research questions such as: ``do we have the best clustering that could be achieved and if not, how can we improve it?'' (usually related to text applications)~\cite{Crossno2009LSAView}. 
  \item[\textcolor{ColT6}{\rule{\boxh}{\boxh}} Topic 6 -- models' explanations \& visualization evaluation.] \quad 
  This topic is rather generic (with 39 papers allocated) as it addresses the \emph{understanding/explanation of ML models}~\cite{Gil2019Towards,Tamagnini2017Interpreting}. For many visualization tools belonging to this topic class, we can observe that \emph{user studies} (i.e., evaluations)~\cite{Brooks2015FeatureInsight,Badam2017Steering,Gotz2016Adaptive,Kapoor2010Interactive,Sun2017Label,Zhang2016AVisual} have been performed with participants from different educational levels (novices, practitioners, ML experts, and so on). Following the overall theme of this STAR, a straightforward unsolved problem in this area is to find answers to how exactly we shall progress with the development of visualization tools for boosting the trust in ML models and their results.
  \item[\textcolor{ColT7}{\rule{\boxh}{\boxh}} Topic 7 -- subspaces exploration \& distances examination.] \quad 
  Clustering and DR are both covered together when exploring subspaces~\cite{Assala2016Interactive,Krause2016SeekAView,Lee2014A}. Finding the correct distance function, checking if these distances are preserved after the projection from the high-dimensional space into the 2D space, and matching the \emph{users' cognitive expectations} is clearly not a trivial task. 
  As a result, many papers are published in this area~\cite{Albuquerque2011Perception,Brown2012Dis,Jackle2017Pattern,Wang2019High}, making this topic class with 30 papers one of the most prominent in our analysis.
  \item[\textcolor{ColT8}{\rule{\boxh}{\boxh}} Topic 8 -- models' predictions \& design prototyping.] \quad 
  Another generic topic class with 18 related papers contains, among others, the subject of ML models' predictions~\cite{Schneider2017Visual,Xu2019EnsembleLens} that has already been seen in Topic 5. The difference between this class and Topic 5 is the focus of its related papers, which is on the instantiation of visualization prototypes with different design choices that should be carefully considered based on previous InfoVis research. As such, updating the current methods with improved versions can lead to enhanced trust of visualizations and \emph{reduce biases}~\cite{Guo2019Visualizing,Lin2018RCLens,Strobelt2019Seq2seq}.
  \item[\textcolor{ColT9}{\rule{\boxh}{\boxh}} Topic 9 -- points, projection space, \& outliers' exploration.] \quad 
  With 68 papers in total, the area of \emph{outlier detection} is prominent in our categorization, see Table~\ref{tab:categorizationLegend}. 
  This category is even confirmed through the topic analysis (with 19 papers assigned) as many techniques work with \emph{outlier detection}~\cite{Bernard2019Visual,Joia2015Uncovering,Rauber2015Interactive}. Another hot topic in the visualization community is the visual analysis of relations between points and dimensions within the various projection spaces~\cite{Aupetit2007Visualizing,Coimbra2016Explaining}.
  \item[\textcolor{ColT10}{\rule{\boxh}{\boxh}} Topic 10 -- neurons' activations.] \quad 
  When visualizing DL techniques, the existing research has tried to address the activation of neurons in NN and their visual representation ~\cite{Hamid2019Visual,Hazarika2019NNVA,Hohman2020Summit}. 
  Different visualization techniques (e.g., 2D saliency/activation maps) have been used to visualize the activations of such neurons in various DL models, especially for image applications~\cite{Alsallakh2018Do}. This topic class consists of 20 papers about visualizing the internal operations of NNs during the training phase. A possible research question in this context is: ``what else can be visualized (for instance, gradients~\cite{Cashman2018RNNbow}) that gives meaning to humans about the learning process of a NN?'' 
\end{description}

\noindent \textbf{Topic embedding.} \quad
The ten-dimensional data space of the topics over all 200 papers has been reduced to two dimensions by using t-SNE~\cite{vanDerMaaten2008Visualizing}, i.e., two papers are positioned close to each other if their topic relationships are alike, see~\autoref{fig:topics_proj}(a). The scales in the depicted bar charts are from 0 to 1, with 1 being the highest relevancy value of a topic in~\autoref{fig:topics_proj}(b) and of a term in~\autoref{fig:topics_proj}(c). The black outlines in the 2D embedding (see \autoref{fig:topics_proj}(a)) were appended manually.

As it can be derived from~\autoref{fig:topics_proj}(b), Topics 6 \&~7 are the most prominent ones, followed by Topics 5 \&~9, 8 \&~10, and the others. In more detail, \emph{ML models' explanation and visualization systems evaluation} (Topic 6) and \emph{subspaces exploration and distances examination in clustering and DR} (Topic 7) are two discussed topics that cover approximately 35\% of all papers. With regard to~\autoref{fig:topics_proj}(c), some interesting top terms are---as expected---``models'' (ML), ``image data'' (computer vision, see even~\autoref{tab:data}), ``layers'' (DL), ``clusters'', ``topic'' (analysis), ``subspace'', ``projections'', and ``dimensions'' (DR). By observing the t-SNE projection in~\autoref{fig:topics_proj}(a), we can find more interesting insights. For instance, the tightest cluster is color-encoded in green and related to \emph{NNs models' hyper-parameters and reward visualization during training for image applications} (Topic 3). Another interesting result is that the misclassification of orange (Topic 2) \& pink (Topic 7) points as well as of pink (Topic 7) \& red (Topic 4) points in the embedding happens due to three concept terms that are spread in all three topics, namely, the terms ``clustering'',  ``dimension'', and ``projections''. Furthermore, as Topic 6 is rather generic (\emph{ML models' explanations}), there are some points laid out in-between (i.e., mixed points) with Topics 1 (\emph{DL}) \&~9 (\emph{projections}). Lastly, Topic 8 (\emph{models' predictions \& design prototyping}) is also rather general, because the points in the projection are spread through two other topics (5 \&~10); and this is probably because NNs are a subclass of ML models and Topic 5 (\emph{models' predictions}) is very similar to Topic 8.

Overall, we notice that the automatically generated topics introduce new subcategories (and ideas) that have been discussed in parallel to our categorization and---in consequence---supported even more the categorization of the papers described in the previous section. For instance, Topics 1 and 10 represent VA tools focusing on the visualization of the NNs hidden states and neurons' activations respectively to facilitate the \emph{understanding/explanation} of them. In addition, Topic 1 covers examples for the \emph{comparison} of models based on the visualization of their parameter spaces. Similarly, Topic 2 is related to the \emph{in situ comparison} of concrete models to investigate different behaviors of the ML models. Topic 3, instead, focuses more on \emph{diagnosing/debugging} the training process for reinforcement learning, and Topic 4 reflects the \emph{comparison of data structures} with the use of projections and DR. The remaining topics are explicitly connected to our TL categorization within the corresponding topic list items above. We believe that this mixture of coarse-grained manual categorization with a fine-grained automatic refinement may help guiding potential readers to more insights and analyze the surveyed papers even further.

\subsection{Correlation and Summarization of Categories}

\begin{figure}[tb]
 \centering 
 \includegraphics[width=\columnwidth]{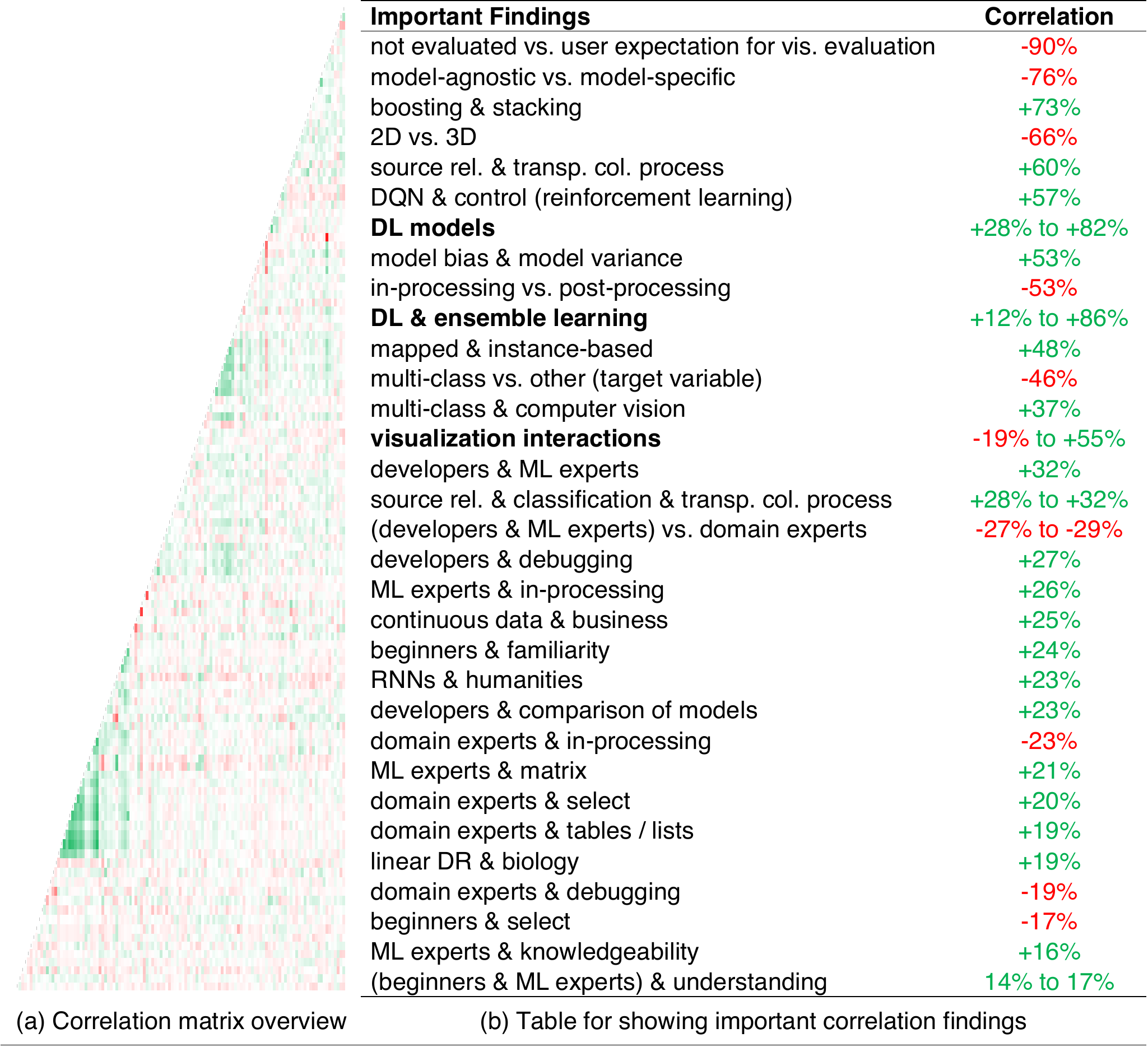}
 \caption{The matrix of correlation values for categories (see \autoref{tab:categorizationLegend}) calculated from our survey data set. In subfigure (a), we show the correlations for the categorization in a thumbnail version of the correlation matrix. The entire correlation matrix for further exploration can be found in \textbf{S7}. (b) To help the reader to see potentially interesting findings, we list the cases in decreasing order of their correlation strengths (independent of negative or positive correlation). In both subfigures, we use red color for negative correlation and green for positive. Cases highlighted in bold refer to groups of categories in general (e.g., visualization interactions).}
 \label{fig:correlation}
 \vspace{-.5em}
\end{figure}

\vspace{.5em}
\noindent \textbf{Correlation between categories.} \quad
We have conducted a correlation analysis for the categories used in our collected survey data set. Individual visualization papers were treated as observations, and categories  (cf. \autoref{tab:categorizationLegend} and \textbf{S5}) were treated as dimensions/variables. Linear correlation analysis was then used to measure the association between pairs of categories. The resulting matrix in~\autoref{fig:correlation} contains Pearson's \emph{r} coefficient values and reveals specific patterns and intriguing cases of positive (green) and negative (red) correlation between categories. Since the interpretation of the coefficient values seems to differ in the literature~\cite{Cohen1988Measurement,Evans1996Straightforward,Taylor1990Interpretation}, we focus on values of correlations that appear interesting to us despite a potentially strong or weak correlation level. Due to the extensive size of the correlation matrix, we include only a thumbnail of it and refer the reader to \textbf{S7} for more detail. In \autoref{fig:correlation}, we present some strong, medium, and weak correlation cases that caught our attention.

The strongest case of \emph{negative} correlation in our data set is the \emph{not evaluated} category $vs.$ \emph{user expectation for evaluation} (cf. 6.6. and 6.7.5.), which clearly highlights the need for further evaluation of visualization tools and techniques. 
Further interesting cases mainly include competing categories from the same group.  For example, \emph{model-agnostic} techniques contradict \emph{model-specific} techniques, because they consider different visualization granularities for a given ML model. \emph{2D} and \emph{3D} oppose each other as typically only one of them exists in a visualization approach. Moreover, techniques that focus on data exploration, explanation, and manipulation related to the \emph{in-processing} phases of an ML pipeline are very different compared to systems that monitor the results in the \emph{post-processing} phase of an ML model. The strong negative correlation between \emph{multi-class} and \emph{other} target variables might point to an effect that comes from our own categorization procedure: when papers could not be mapped to a concrete target variable (\emph{multi-class}, for instance), then the \emph{other} category has been assigned, e.g., to show the irrelevance of the target variable for a visualization technique. The category \emph{domain experts} is negatively correlated to managing models during the \emph{in-processing} ML phase, which makes sense as they do often not know much about how models work. Similarly, \emph{developers} and \emph{ML experts} together are weakly but negatively correlated with \emph{domain experts} confirming the previous acquisition. Other insights are that \emph{beginners} are not usually using \emph{selection} as interaction technique and \emph{domain experts} do not work with \emph{diagnosing/debugging} ML models as they do not have the experience and/or knowledge following the previous inference.

\begin{figure*}[ht!]
 \centering 
 \includegraphics[height=.95\textheight]{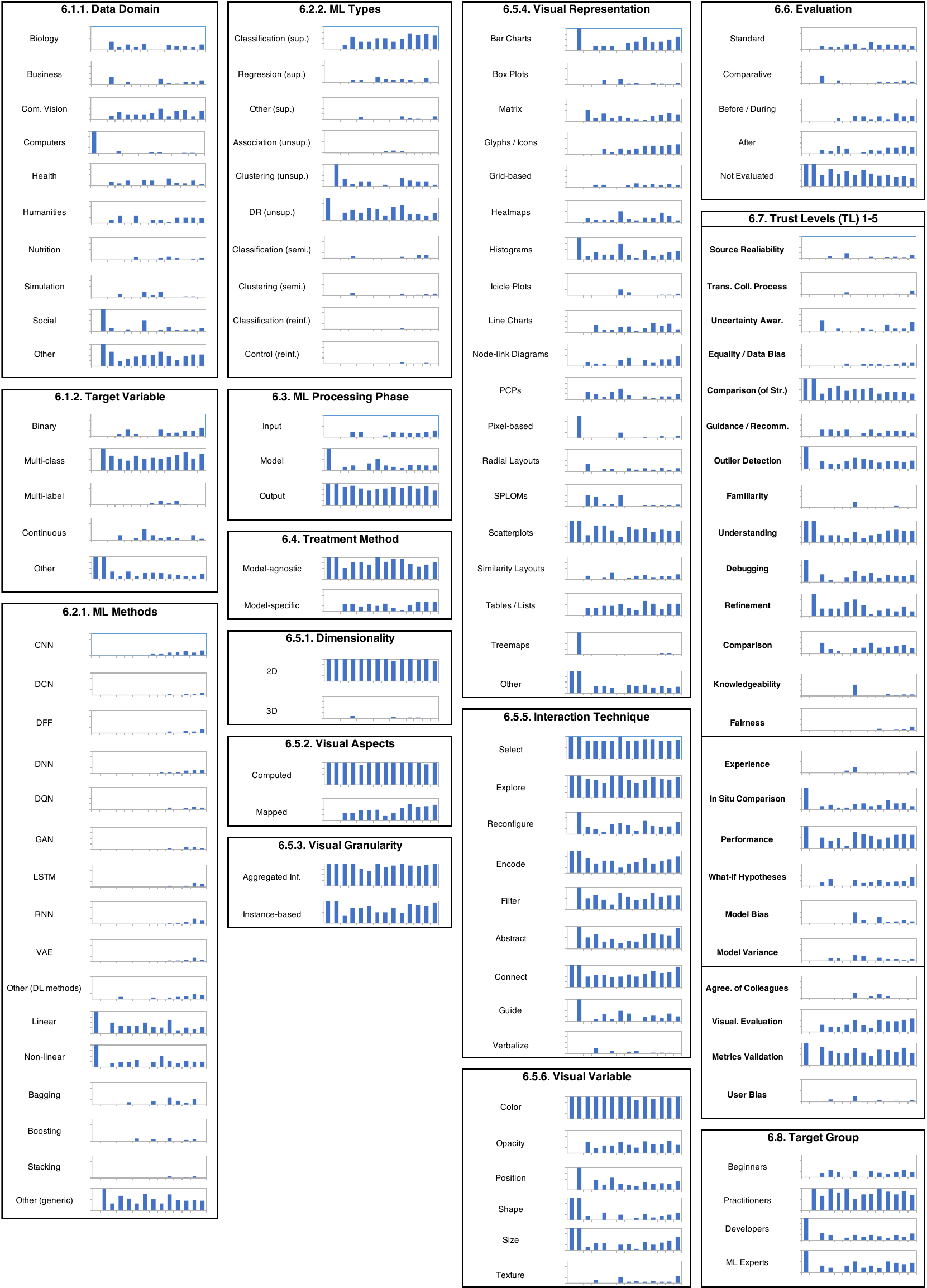}
 \caption{Sparkline representation of relative category popularity over time during 2007--2020. Each bar represents the support for a specific category (cf. \autoref{tab:categorizationLegend}) in relation to the total count of techniques in the same year in our data set.}
 \label{fig:temporal}
 \vspace{-.9em}
\end{figure*}

Cases with \emph{positive} correlation start in \autoref{fig:correlation}(b) with \emph{stacking} which is highly correlated with \emph{boosting} ensemble learning as the former sometimes includes the latter technique. All \emph{DL techniques} among each other have on average a medium positive correlation, which shows that they have much more in common compared to other ML methods, for example, DR. The same is true for the group of \emph{visualization interactions} to a slightly less extend. When \emph{source reliability} is taken into account and researched by scientists, then the \emph{transparent collection process} is usually examined together. \emph{Deep Q-networks (DQNs)} are positively correlated to and seem to be normally used together with \emph{reinforcement learning} methods, particularly with the subtype \emph{control}. Furthermore, when \emph{model bias} challenges are addressed by visualization, then \emph{model variance} is another category that is addressed simultaneously. 
\emph{Ensemble learning} along with \emph{DL} are positively correlated, as the former includes the latter in many cases (a fact already mentioned before). \emph{Mapped instances} lead to \emph{instance-based} visualizations, in general. With regard to domains, \emph{continuous} target variables and \emph{business} are positively correlated (potentially due to trend predictions); as well as \emph{computer vision} and \emph{multi-class} data in an even stronger fashion. The latter correlation is also supported by our data set analysis (see \autoref{tab:data}). Finally, \emph{visualization interactions} are positively correlated among each other, with the exception of \emph{verbalization} which is negatively correlated with the remaining categories in this group. This possibly means that \emph{verbalization} is not frequently used by the VA system developers. For more details, we refer the reader to \autoref{fig:correlation}(b) and supplementary material \textbf{S7}.

\vspace{.5em}
\noindent \textbf{Popular approaches.} \quad
The statistics in \autoref{tab:categorizationLegend} support our expectations of the most common aspects of existing visualization techniques for enhancing the trustworthiness of ML models.
For our first aspect (6.1. Data), \emph{computer vision}, \emph{humanities}, \emph{health}, and \emph{biology} seem the most prominent \emph{domains} in the surveyed papers. \emph{Multi-class classification} is the most common target variable in our discussed techniques. Furthermore (6.2. ML), \emph{linear} and then \emph{non-linear DR} techniques are commonly used, followed by \emph{bagging (ensemble learning)} and \emph{CNNs} from the \emph{DL} class. The vast majority of the papers address \emph{supervised learning} and specifically \emph{classification} problems, and in second position \emph{DR} and \emph{clustering} which belong to \emph{unsupervised learning}. (6.3. and 6.4.) \emph{Post-processing} and \emph{model-agnostic} visualization techniques cover around 75\% of all papers. (6.5.) With regard to \emph{visual aspects} and \emph{granularity}, almost all techniques used have at least a component which is \emph{computed} and not \emph{mapped/derived} from the data directly; and \emph{aggregated information} is slightly more common than \emph{instance-based/individual} exploration of instances. 

The absolute majority of the visualizations rely only on \emph{2D representations}, and \emph{color} is the visual channel most commonly used for encoding information in the corresponding visualization systems, tools, and techniques. 
The rather large number of techniques using \emph{opacity} to hide points/instances and \emph{size/area} to encode data attributes can be explained by the extensive usage of \emph{scatterplots}. 
Other popular visualizations are \emph{bar charts}, \emph{custom glyphs} and \emph{specialized icons}, \emph{histograms}, and finally, \emph{line charts}. 
More traditional visual representations, such as \emph{tables}, \emph{lists}, and \emph{matrices}, are working in pairs with \emph{instance-based} exploration techniques, which are far less complicated than the previously mentioned visualizations. On the interaction side, \emph{selection}, \emph{exploration}, and \emph{abstraction/elaboration} are the three most prominent categories found in many papers, followed by other interaction techniques, such as \emph{connecting} all the different views, \emph{filtering out} or \emph{searching} for specific instances, and \emph{encoding}.
(6.6.) Around half of the visualization techniques that we analyzed have \emph{not been evaluated}. 

The \emph{trust levels} (6.7.) show that more works tackle \emph{source reliability} problems rather than the \emph{transparent collection process} challenge (as seen in TL1). For the second level (TL2), researchers focus on the \emph{comparison of structures} and \emph{outlier detection}. In the third level (TL3), \emph{understanding}, \emph{steering}, \emph{comparing}, and \emph{debugging} ML methods are quite popular. These aforementioned categories can be considered under the umbrella of interpretable/explainable ML methods. For TL4, \emph{performance}, \emph{in situ comparison}, and \emph{what-if hypotheses} are other very often occurring categories connected with the selection process of an individual ML model. Ultimately in TL5, \emph{metrics validation and results observation} at the final stage of the processing phase is the most frequent category with 130 papers. Last but not least for 6.8., the visualization systems and techniques have as a main \emph{target group}, usually \emph{practitioners/domain experts}, followed by \emph{ML experts} with large distance. 
The analysis above sheds light into the reasons why a few approaches seem to be more popular than others. The ML side uses mostly performance, metrics validation, and results to monitor and boost trust in the ML models.
In contrast, the visualization side focuses more on traditional visual representations and/or multivariate, scalable visualizations that the experts are more willing to use.

\vspace{.5em}
\noindent \textbf{Temporal trends.} \quad 
While the analyses presented above focus on the overall statistics, we have also analyzed the temporal trends for individual categories based on the collected data. 
\autoref{fig:temporal} provides a sparkline-style representation of the information about each category's support (i.e., the count of corresponding techniques) over time. 
The values are normalized by the total count of techniques for each respective year between 2007 and 2020 (for example, 3 out of 9 papers from 2010 used \emph{computer vision data} to demonstrate the usability of their tools). 
The resulting representation in \autoref{fig:temporal} allows us to confirm, for instance, that the \emph{ML processing phase} visualized consistently most often is \emph{post-processing} rather than \emph{in-processing} or \emph{pre-processing}. 
Combination of such temporal trends with the overall statistics also allows us to identify and further discuss the usage of currently underrepresented categories.

\vspace{.5em}
\noindent \textbf{Underrepresented categories.} \quad 
\emph{Multi-label data} and \emph{computer-related data} (from software or hardware) are two underrepresented categories that show no trend for a potential increase according to \autoref{fig:temporal}. For \emph{ML methods}, approaches such as \emph{stacking ensemble learning}, \emph{deep convolutional networks (DCNs)}, and \emph{DQNs} are also not covered in detail. Nevertheless, there is a very small increasing trend for them observable in \autoref{fig:temporal}.
Explicit techniques addressing problems that come with \emph{stacking ensemble learning} were not found in any paper, thus indicating a new research opportunity.
For \emph{ML types}, the subcategory of solving \emph{classification} problems while using \emph{reinforcement learning} is almost never visualized and actually never addressed explicitly by the visualization community. 
Other underrepresented categories here are \emph{reinforcement learning} and \emph{control}, and \emph{association} for \emph{unsupervised learning}. 

For the \emph{visual representation}, the \emph{treemaps} and \emph{icicle plots} categories are virtually not supported by the data. Further techniques that belong to the last category within \emph{visual representation} (``\emph{other}'') and are fairly underrepresented are \emph{waterfall charts}, \emph{bipartite visualizations}, and lastly \emph{area charts} (as also mentioned in~\autoref{sec:categ}). For the \emph{interaction techniques}, the category of \emph{verbalization} emerged in 2010 and has not attracted much support in the publications; even though recently in 2018, Sevastjanova et al.~\cite{Sevastjanova2018Going} argued about the importance of its existence. Moreover, \emph{texture} is the least usual way to represent the data visually in comparison to the others. \emph{Comparative evaluations} are the rarest way of \emph{evaluating visualizations}, which is rather logical because not every technique has an obvious opposing one. 

The real challenges start when we check the \emph{trust levels} aspect because many techniques are underrepresented, which means there are several research opportunities in the area. \emph{Transparent collection processes}, \emph{source reliability}, and \emph{equality/data bias} are usually not covered by papers. Other problems, such as how visualization can assist with the \emph{familiarity} a user has for a \emph{learning method}, should also be in the research agenda of our community. \emph{Fairness} (and previously mentioned \emph{equality} for the data) of the \emph{learning methods} seem to be in the spotlight according to the temporal statistics (see \autoref{fig:temporal}). Finally, \emph{developers} (i.e., model builders) and \emph{beginners} are the two most underrepresented target groups in the papers we analyzed. \emph{Knowledgeability} about \emph{learning methods} and details available to different types of users is not well supported. As a result, customization and reconfiguration of visualizations that take into account the \emph{experience} of users in order to choose a specific ML model are not researched to the required extent. Furthermore, a few techniques enable \emph{agreement of colleagues} and study about the consequences of using provenance in visualization tools in order to cover our discussed subject. \emph{User bias} is ignored in almost all of the visual systems. 

All of the underrepresented categories discussed above might be candidates for open challenges, as can be seen in Section~\ref{sec:research}. From an ML perspective, the most real-world challenges are about either classification or regression problems. Consequently, other ML types are not researched to the same level. From the visualization perspective, a large amount of time and effort is necessary to design and perform a ``proper" visualization evaluation~\cite{Lucke2018Lowering}. Moreover, as long as the visualization tools do not focus on beginners, familiarity with and knowledgeability of the algorithms are left aside by visualization researchers. 

\begin{table*}[ht!]
 \centering 
 \caption{Overview of data sets ordered according to their usage. A ``\#'' in the columns indicates the number of papers in this survey using a particular data set; the data sets are then further grouped based on this number.}
 \includegraphics[width=\textwidth]{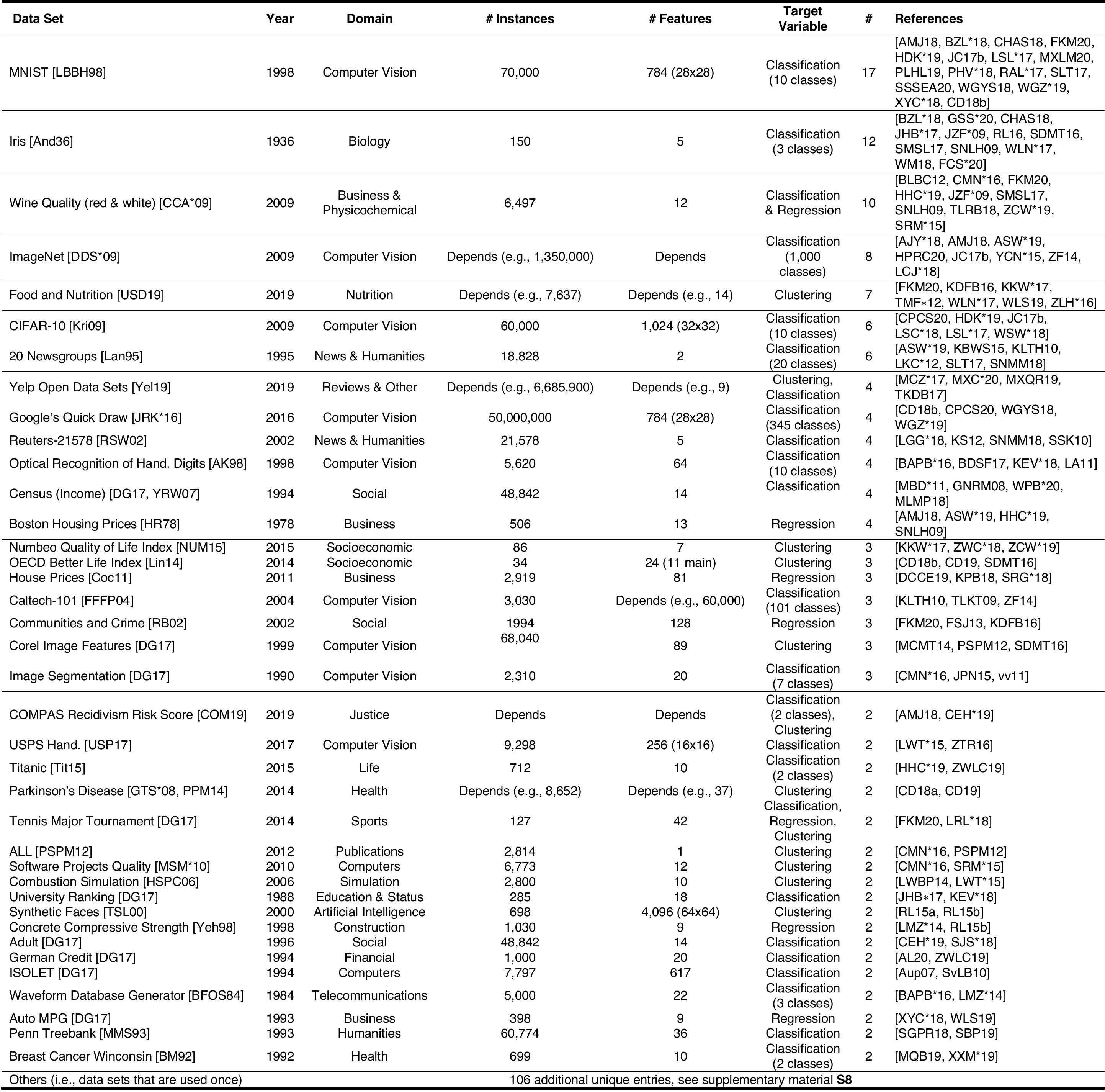}
 \label{tab:data}
 \vspace{-.5em}
\end{table*}

\begin{figure*}[ht!]
 \centering 
 \includegraphics[width=\textwidth]{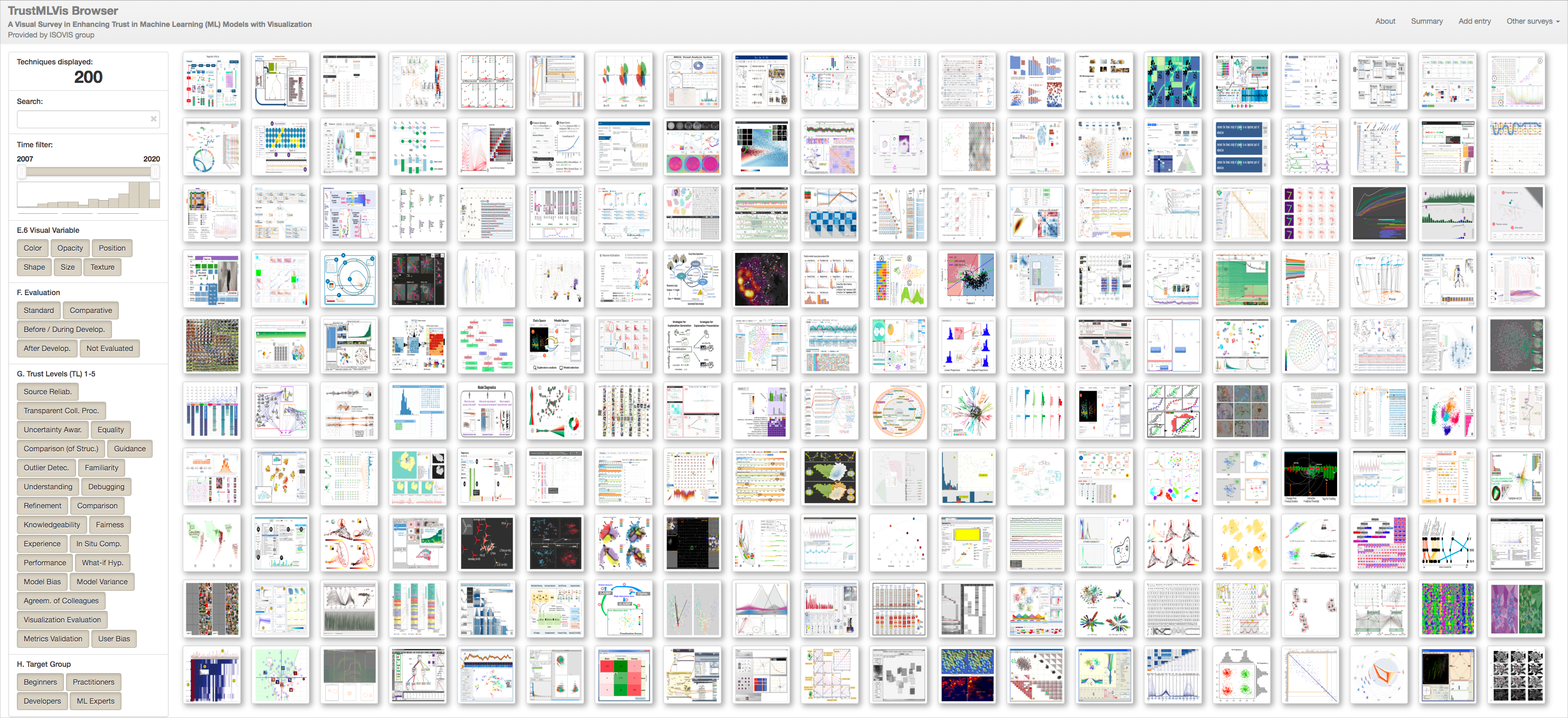}
 \caption{The user interface of TrustMLVis Browser (available at \href{https://trustmlvis.lnu.se}{trustmlvis.lnu.se}), an interactive survey browser accompanying this article.}
 \label{fig:web}
 \vspace{-.9em}
\end{figure*}

\subsection{Data Set Analysis}
\noindent \textbf{Methodology.} \quad
For the data set analysis, we consider only non-synthetic (i.e., not artificial) data sets which can be accessed online. We also include data sets that can be requested from the paper authors. For the individual data features, we take further into account the labels (i.e., classes), if they are existent. Overall, details about a data set were collected relying on the description provided by the authors of a paper, for example, how they collected and stored the data. In any other case, we omitted the data sets. All data sets are sorted first according to the number of occurrences in the 200 papers, and then by year to show the most recent first.

\noindent \textbf{Results.} \quad
The result of this process can be observed in \autoref{tab:data}. In the listed 38 cases, the data sets are used in at least two papers, and the remaining 106 entries are used once only (cf. \textbf{S8}). In total, we managed to identify 144 non-synthetic data sets
%
~\nocite{Alimoglu1997Combining,Alpaydin1998Cascaded,Anderson1936The,Argenziano2000Interactive,Arrhythmia2005,Becker2016Identifying,Belhumeur1997Eigenfaces,Bennett92robustlinear,BestCity2012,BPD2016,Breiman1984Classification,Broemstrup2010Molecular,Bruni2014Multimodal,Cadaster2009,Cettolo2012Wit,Chang2008Importance,Chase1992The,Cock2011Ames,Cohen2006Reducing,COMPAS2019,Cortez2007A,Cortez2009Modeling,Craven1998Learning,Cresci2017The,Deng2009ImageNet,Dua2017UCI,Duff1989Sparse,Airfoil1998,Dor1987Attributes,Elmqvist2008Rolling,English2018Football,Ess2008A,EuropeanESS2018,Fanaee2014Event,Fei2004Learning,Fernandes2015A,Finkelstein2001Placing,Freire2009Short,Friedman2002Stochastic,Gabrilovich2004Text,Gansner2005Topological,Goetz2008Movement,Greene2005Producing,Grgic2011SCface,Griffin2007Caltech,Harisson1978,Hawkes2006Direct,Henderson1981Building,Hickey2014A,Higuera2015,Hill2015SimLex,Huuskonen2000Estimation,Hyperspectral2019,ImageCLEF2019,InfoVis2017,jassby2017WQ,Jongejan2016The,Kemp2000Analysis,Kim2014A,Kononenko1984Experiments,Krizhevsky2009Learning,Lang1995NewsWeeder,Lecun1998Gradient,Lendasse2007Variable,Li2002Learning,Lids2019,Lind2014Better,Liu2015Deep,Maggiori2017Can,Mandelli2013Scenario,Mansouri2013Quantitative,Marcus1993Building,Marti2002The,Matthaus2009Interactive,MAWI2019,Meirelles2010A,Mikolov2013Linguistic,Moro2014A,Munzner2011Reflections,Nam2013Trip,Nene1996Columbia,NewYork2017,Nilsback2006A,NIP2017Adversarial,Norman2008Contest,NUMBEO2015,ODP2019,Olusola2010D,OpenML2014,Otto2014,Over2003An,Paiva2012Semi,Pang2005Seeing,Pang2008Opinion,Parkinson2014,Paulovich2008Least,Pozzolo2015Calibrating,Redmond2002A,Revow1996Ringnorm,Rose2002The,Samet2000The,Scotch2017,SDO2019,Simonoff2012Smoothing,Smarr2016Detection,Smith1988Using,Smith2006Computational,Spooky2017,Tang2008ArnetMiner,Tenenbaum2000A,Tetko2001Estimation,Titanic2015,TransitFeeds2019,Tron2007A,USDA2019,USPS2017,Putten2000challenge,vanUnen2016Mass,Veta2015Assessment,World2019,Wu2010A,Xiao2017Fashion,Xu2017ViDX,Yan2012Generalized,Yang2007Analysis,Yeh1998Modeling,Yeh2009The,Yelp2019,Zelenyuk2005Single} 
in our 200 surveyed papers. The most frequently occurred data sets are MNIST~\cite{Lecun1998Gradient}, Iris~\cite{Anderson1936The}, Wine Quality~\cite{Cortez2009Modeling}, ImageNet~\cite{Deng2009ImageNet}, Food and Nutrition~\cite{USDA2019}, CIFAR-10~\cite{Krizhevsky2009Learning}, and 20 Newsgroups~\cite{Lang1995NewsWeeder}. 3 out of these 7 data sets are about computer vision and are usually used in papers that work with DL and NNs. Validating our previous categorization, \emph{classification} and then \emph{clustering} problems are the more occuring \emph{target variables}, and finally \emph{regression}. The number of instances and features can be found in our table along with the number of classes for some cases (if available). The individual papers that used the data sets are listed in the rightmost column of \autoref{tab:data}; and references to the data set providers are given together with the name of the data sets in the first column.

\section{Discussion and Research Opportunities} \label{sec:discuss}
In this section, we discuss our online survey browser. Afterwards, we move on to research opportunities based on the data-driven analyses presented in~\autoref{sec:topic}.

\subsection{Interactive exploration with a survey browser}

Our work on this survey has been complemented by the development of an interactive survey browser~\cite{Beck2016Visual,Kucher2014Text,Kucher2015Text,Kucher2018The,Schulz2011TreeVis,Tominski2013The} similar to our group's previous contributions on text and sentiment visualization. 
TrustMLVis Browser is available as a web application, and its user interface (see \autoref{fig:web}) comprises (1) a grid of thumbnails representing visualization techniques and (2) an interaction panel supporting category-, time-, and text-based filtering. 
The user can access the details and bibliographic information about a specific technique by clicking on the corresponding thumbnail. 
Several dialogs with the overall statistics for the complete data set (cf. \autoref{tab:categorizationLegend}) and the supplementary materials are available via the links at the top of the web page. 
We encourage the readers of this article to explore the data with the survey browser and to suggest further candidate entries by using the corresponding \cit{Add entry} dialog.

\subsection{Research Opportunities} \label{sec:research}
\noindent \textbf{The impact of bias.} \quad
By looking at our categorization, we can infer that some level of \emph{bias} might be represented in all our defined trust levels in different forms: (a) \emph{data bias (equality)}, (b) \emph{previous familiarity with algorithms}, (c) \emph{model bias}, and (d) \emph{user bias}. 
Also, it is known that visualization techniques ordinarily do not scale very well when analyzing massive volumes of data. Furthermore, some of the ML approaches have inherent challenges to face, for example, the curse of dimensionality~\cite{Bellman2003Dynamic} in case of DR. 
Thus, considerable levels of selection bias might be unintentionally ignored by the user, for instance, when users have to choose from a selection while not seeing the entire picture and/or the alternatives~\cite{Gotz2016Adaptive,Lespinats2011CheckViz}. 
Hence, the research question here is: ``what novel solutions can help users to minimize the impact of bias with regard to the data?'' 
A potential answer would be to consider various interaction logs with the VA system. Data generated as part of the analysis process could be considered as well. 
This data together with the logs could be processed automatically with additional independent ML models and potentially guide users to improvements of the underlying ML models used in the data analysis process. Hence, the ways of combining automatic methods with smart visualizations~\cite{Shneiderman2020Human} are still not revealed and should be further evaluated with empirical studies as well as quantitative and qualitative experiments.

\noindent \textbf{Alternatives and combination.} \quad
Visualization is often used as the medium enabling human-computer interactions (HCI). 
It usually encourages the development and application of multi-disciplinary methods originating from different areas of research. 
To find an equilibrium state between human and computer controlling the ML process is not a trivial task~\cite{Shneiderman2020Human}. 
Researchers that are intimate with ML models and visualizations are capable of appropriately promoting the joint development of visual explanations for ML models.
Furthermore, there is a possibility to employ verbalization (as discussed before) as a complementary tool alongside visualization for explaining ML models. The challenges of developing visualization systems involving such text explanations and finding the right balance between these two approaches are still open~\cite{Sevastjanova2018Going}. Here, we foresee an open research challenge upon how to combine visualizations, verbalization (text explanations), and voice commands (AI assistants) that should together perform overlapping tasks in complex visualization systems and propose task solutions to the users. 
As can be seen with our categorization, analysts usually deal with data manipulation problems which can lead to compromising the trust, for example, (1) comparison of structures, (2) guidance in data selection, (3) outlier detection, (4) comparison of algorithms, and (5) in-situ comparison of concrete model structures. The aforementioned methods might provide a possible remedy for such compromises of trust. 

\noindent \textbf{Security vulnerabilities.} \quad
When research is conducted in ML, there is always a factor that is not often taken into account at first: ``how do we secure ML models from unethical attacks?'' An instance of this idea is published by Ma et al.~\cite{Ma2019Explaining}, explaining how visualization can assist in avoiding vulnerabilities of adversarial attacks in ML. Specifically, their focus is on how to avoid data poisoning attacks from the models, data instances, features, and local structures perspectives with the use of their VA approach. 
Nowadays, visualization systems are deployed online for users to access them easily. Such internet-accessibility leads to further problems concerning security vulnerabilities. 
This is one of the advantages of TensorFlow.js, which utilizes the WebGL-accelerated implementation of JavaScript in web browsers to implement and use ML models on local computers. 

\noindent \textbf{Fairness of the decisions.} \quad
Going beyond interpretability towards more explainability is another open challenge. However, general proposals of frameworks in the visualization community combining ML and visualizations have been already described in recent research papers~\cite{Ming2019Interpretable,Spinner2019explAIner}. These global frameworks were divided into smaller parts by other works that compare DL methods, for instance~\cite{Murugesan2019DeepCompare}. Further tools explore local trends instead of global patterns~\cite{Zhao2014LoVis}. 
Two further open questions reaching beyond interpretability and explainability are \cit{how fair were all those decisions and what if we have chosen another path?} and \cit{how can fairness be translated between the trust levels?} (cf. the work by Ahn and Lin~\cite{Ahn2019FairSight}).

\noindent \textbf{Ways of communication and collaboration.} \quad
Increasing the users' trust in ML models is not a trivial task. Visualization can assist in this challenge in multiple ways. A good starting point is employing simple techniques, such as querying specific data instances and areas of interest, in a user-friendly way~\cite{Hoferlin2012Inter}. However, the issue of improving trustworthiness in ML with visualization is also related to the issue of improving the trust for visualization itself~\cite{Boy2014A,Borner2019Data}. To achieve the best outcome when evaluating visualization designs, the input data, the goals, and the target group of a visualization should be under the spotlight. On the optimistic side, many papers exist that try to tackle the challenges of evaluation and design choices for visualizations~\cite{Federico2016A,Knudsen2016Using,Kosara2016An,Lucke2018Lowering,Mayr2016Looking,Qu2016Evaluating}.
Development of further guidelines and best practices for (1) how people within different scientific fields and varying backgrounds and experiences should communicate, and (2) which visualization techniques and systems should be established as a standardized interaction medium between them, present another open challenge. As previously discussed, Jentner et al.~\cite{Jentner2018Minions} suggest that metaphorical narratives can explain the ML models to various target groups in a user-friendly way, but further research is required in this regard. 

\noindent \textbf{Almost unexplored areas.} \quad
Related to the non-trust level classes (which implicitly influence trust), we believe that all underrepresented categories can pose as new ideas for novel research. For example, visualization researchers have still not provided sufficient support for some specific NNs, such as convolutional deep belief networks (CDBNs), deep residual networks (DRNs), and multi-column DNNs (MCDNNs). Also in ensemble learning, visualization tools that target solely the boosting techniques are quite rare, e.g., gradient boosting and adaptive boosting (AdaBoost) appear not to be covered to the same level as random forests. 
Another example of such a category is stacking ensemble learning, i.e., constructing a combination (a stack) of different models that should become the input for other meta-model(s). 
Employing visualization to facilitate the experts in developing and using such stacks in a trustworthy way without resorting to trial and error is also an open research challenge. 
Additionally, regression problems are also far less covered than classification. In unsupervised learning, association/pattern mining is uncommonly investigated by visual tools. To conclude this paragraph, reinforcement learning approaches are almost ignored with only a few available papers covering this area of how visualization can help to monitor an automatically controlled learning process~\cite{Saldanha2019ReLVis}. In reinforcement learning, classification tasks, i.e., letting an agent act on the inputs and learn value functions~\cite{Wiering2011Reinforcement}, are not once addressed by visualization.

\section{Conclusion} \label{sec:conclusion}
	In this survey, we study the state of the art in enhancing trust in machine learning (ML) models with the use of visualizations. 
We introduced the background necessary for defining trustworthiness of ML models and explained the methodology used to select relevant papers in the literature. 
Based on the selected 200 peer-reviewed publications that introduce a large variety of visualization techniques to increase trust in ML models and their results, we proposed a fine-grained categorization comprising 8 high-level aspects partitioned into 18 category groups that on their part contain 119 categories in total. 
In addition, we performed a topic analysis to be able to discover connections and emerging topics among the 200 papers. 
Further analyses of the categorized data involved category correlations, temporal trends, and data sets used in the respective publications. 
In order to make our categorization and the assignment of papers into categories accessible for the public, an interactive survey browser---called TrustMLVis Browser---was implemented and made available online. 
It supports the readers of this STAR in the exploration of the rich information provided in this work, thus facilitating future research in enhancing trustworthiness of ML models with the help of interactive visualizations.  
Our findings indicate the growing interest for developing visualizations in ML to improve trustworthiness in the context of various data domains, tasks, and multidisciplinary applications. As future work, we intend to continue extending and refining the survey data set, categorization, and corresponding analyses, as well as maintaining the online survey browser.

\bibliographystyle{eg-alpha-doi}

\bibliography{STAR_STAR_EuroVis2020}

\end{document}